\documentclass[conference]{IEEEtran}
\IEEEoverridecommandlockouts
\usepackage{cite}
\usepackage{amsmath,amssymb,amsfonts}
\usepackage{algorithmic}
\usepackage{graphicx}
\usepackage{textcomp}
\usepackage{xcolor}
\usepackage{subcaption}
\usepackage{hyperref}
\def\BibTeX{{\rm B\kern-.05em{\sc i\kern-.025em b}\kern-.08em
    T\kern-.1667em\lower.7ex\hbox{E}\kern-.125emX}}

\begin{document}

\title{Multi-Strategy Enhanced COA for Path Planning in Autonomous Navigation}

\author{
\IEEEauthorblockN{
    Yifei Wang\textsuperscript{1}, Jacky Keung\textsuperscript{1}, Haohan Xu\textsuperscript{2}, 
    Yuchen Cao\textsuperscript{1}, Zhenyu Mao\textsuperscript{1,*}
}
\IEEEauthorblockA{\textsuperscript{1}Department of Computer Science, City University of Hong Kong, Hong Kong, China\\
    ywang4748-c@my.cityu.edu.hk, Jacky.Keung@cityu.edu.hk, cynthcao2-c@my.cityu.edu.hk, zhenyumao2-c@my.cityu.edu.hk}
\IEEEauthorblockA{\textsuperscript{2}IT Department of Materials Company, PetroChina Southwest Oil \& Gasfield Company, Chengdu, China\\
    haohanxu@petrochina.com.cn}
\normalsize *corresponding author
}

\maketitle

\begin{abstract}

Autonomous navigation is reshaping various domains in people's life by enabling efficient and safe movement in complex environments.
Reliable navigation requires algorithmic approaches that compute optimal or near-optimal trajectories while satisfying task-specific constraints and ensuring obstacle avoidance.
However, existing methods struggle with slow convergence and suboptimal solutions, particularly in complex environments, limiting their real-world applicability.
To address these limitations, this paper presents the \textit{M}ulti-Strategy Enhanced \textit{C}rayfish \textit{O}ptimization \textit{A}lgorithm (MCOA), a novel approach integrating three key strategies: 1) Refractive Opposition Learning, enhancing population diversity and global exploration, 2) Stochastic Centroid-Guided Exploration, balancing global and local search to prevent premature convergence, and 3) Adaptive Competition-Based Selection, dynamically adjusting selection pressure for faster convergence and improved solution quality.
Empirical evaluations underscore the remarkable planning speed and the amazing solution quality of MCOA in both 3D Unmanned Aerial Vehicle (UAV) and 2D mobile robot path planning.
Against 11 baseline algorithms, MCOA achieved a 69.2\% reduction in computational time and a 16.7\% improvement in minimizing overall path cost in 3D UAV scenarios.
Furthermore, in 2D path planning, MCOA outperformed baseline approaches by 44\% on average, with an impressive 75.6\% advantage in the largest 60*60 grid setting.
These findings validate MCOA as a powerful tool for optimizing autonomous navigation in complex environments. The source code is available at: \href{https://github.com/coedv-hub/MCOA}{\textcolor{blue}{https://github.com/coedv-hub/MCOA}}.

\end{abstract}


\section{Introduction}
\label{sec:introduction}

With rapid advancements in artificial intelligence, sensor technology, and computational capabilities, autonomous navigation have been widely adopted across various domains \cite{nahavandi2022comprehensive}.
For systems operating in dynamic, unstructured, or highly structured environments, such as UAVs, mobile robots, and autonomous vehicles, autonomous navigation is essential for enabling independent perception, decision-making, and task execution. 
At the core of autonomous navigation lies path planning, a critical process that ensures safe, efficient, and adaptable movement. 
As a critical process, path planning focuses on computing an optimal or near-optimal trajectory that satisfies task-specific constraints while ensuring obstacle avoidance \cite{liu2023path}.  


Path-planning algorithms and optimization mechanisms have seen significant advancements, enabling autonomous navigation across diverse operational scenarios.
Heuristic and metaheuristic algorithms, such as Genetic Algorithm (GA) \cite{mirjalili2019genetic}, Particle Swarm Optimization (PSO) \cite{kennedy1995particle}, and Ant Colony Optimization (ACO) \cite{dorigo2007ant}, have proven effective in solving complex path-planning problems through iterative optimization \cite{abdel2018metaheuristic}.
Building on these algorithms, recent advancements in multi-strategy optimization mechanisms, including multi-point leadership crossover and adaptive competitive swarm optimization, have demonstrated promising results in addressing multi-objective optimization problems (MOPs) \cite{zhang2023salp, huang2022multi}.

However, existing methods often struggle with premature convergence and suboptimal path solutions.
Such solutions are frequently characterized by unnecessary detours, excessive computational costs, and poor adaptability in dynamic environments \cite{liu2023path}.
Moreover, the slow convergence rate of iterative optimization methods further limits their applicability in real-time path-planning scenarios \cite{liao2021f}.
These challenges emphasize the persistent trade-off between global exploration and local exploitation, a critical barrier in achieving robust and efficient path-planning solutions.



This paper presents MCOA, a novel approach designed to overcome existing limitations by mitigating premature convergence and improving the solution quality in path planning in complex autonomous navigation environments.  
The key innovations of this work include:
\begin{itemize}
    \item A refractive opposition-based population selection strategy that enhances population diversity, strengthening global exploration and reducing the risk of premature convergence.
    \item A stochastic centroid-guided exploration strategy that effectively balances local and global search, increasing adaptability and minimizing entrapment in local optima.
    \item An adaptive competition-based cave selection strategy that dynamically regulates selection pressure, promoting faster convergence and more efficient optimization.
\end{itemize}


The structure of this paper is as follows: Section \ref{sec:background} provides a concise overview of the research background. Section \ref{sec:proposal} details the proposed MCOA. Section \ref{sec:evaluation} evaluates MCOA in both 3D UAV and 2D mobile robot path planning scenarios. Finally, Section \ref{sec:conclusion} summarizes the study and introduces directions for future research.
\section{Background and Related Works}
\label{sec:background}

\subsection{Autonomous navigation and path planning}

Autonomous navigation is a key technology that empowers unmanned systems to achieve independent, safe, and efficient movement in unknown or dynamic environments \cite{shit2020precise}, with applications customized for various platforms, including UAVs and mobile robots as primary examples.
UAV systems emphasize advancements in spatial navigation, enabling dynamic target tracking in military reconnaissance \cite{gargalakos2024role}, precise agricultural field inspection with centimeter-level positioning \cite{velusamy2021unmanned}, and disaster rescue operations in complex terrains through obstacle avoidance algorithms \cite{li2021development}. 
Mobile robots focus on structured environment interaction, utilizing Simultaneous Localization and Mapping (SLAM) technology for intelligent material transfer in factories \cite{liu2022robot}, integrating indoor positioning systems for hospital pharmaceutical delivery \cite{kyrarini2021survey}, and building autonomous operational loops for home cleaning through environmental modeling \cite{holland2021service}.
Reliable autonomous navigation relies on three core technologies. First, multi-sensor fusion enables dynamic obstacle recognition and semantic map construction, enhancing environmental perception. Second, kinematic modeling defines platform-specific physical constraints and motion characteristics. Finally, path planning integrates these components to dynamically optimize navigation strategies, ensuring safe and efficient movement \cite{xie2023research}.

Path planning enables autonomous systems to determine a collision-free and efficient route from an initial position to a target destination within a given environment \cite{hu2018dynamic}.
To achieve this, a variety of algorithmic methods have been devised to calculate optimal or near-optimal paths that satisfy task-specific constraints while ensuring avoiding obstacle.
Deterministic algorithms such as A* \cite{astar} and Dijkstra’s algorithm \cite{dijkstra} provide reliable solutions but struggle with scalability and adaptability in dynamic environments \cite{wenzheng2019improved}.
Metaheuristic approaches, including GA, PSO, and ACO, have been explored for adaptive and probabilistic path planning \cite{yahia2023path}.
However, these methods often suffer from inefficiencies in search performance, real-time adaptability, and maintaining a balance between exploration and exploitation, necessitating the need for more advanced optimization techniques \cite{liu2023path, liao2021f, aggarwal2020path}.

\subsection{Crayfish optimization algorithm}

Crayfish Optimization Algorithm (COA) was initially proposed by Heming Jia et al. \cite{jia2023crayfish}, inspired by the behavioral characteristics of crayfish, a highly adaptable freshwater crustacean. 
Crayfish thrive in humus-rich sediment, utilizing it for feeding and hiding, and exhibit temperature-dependent behaviors: they seek shelter in caves during high temperatures and forage actively within their preferred range of 20-30°C. 
Based on these adaptive behaviors, the authors abstracted three algorithmic stages: summer resort, competition, and foraging, integrating them into the exploration and exploitation phases in the algorithm, as described below.

\subsubsection{Exploration phase}

When crayfish search for burrows, competition occurs randomly. In COA, this is modeled with $rand<0.5$, and the crayfish position is updated using Eq.\ref{eq1}.
\begin{equation}
    X^{t+1}_{i,j}=X^t_{i,j}+C_2\times rand\times(X_{shade}-X^t_{i,j})
    \label{eq1}
\end{equation}
Here, $X^{t+1}_{i,j}$ is the crayfish position at iteration $t+1$; $X_{shade}$ is the cave location calculated from the current and global optimal solutions, $X^t_{i,j}$ is the position at iteration $t$; and $C_2$ is an adaptive parameter computed by Eq.\ref{eq2}.
\begin{equation}
    C_2=2-(\frac{t}{T})
    \label{eq2}
\end{equation}
where $t$ is the current iteration number and $T$ is the maximum number of iterations.

\subsubsection{Exploitation phase}

\begin{itemize}
    \item Competition case stage
\end{itemize}

When $temp>30$ and $rand\geq0.5$, crayfish enter the competition for burrows stage, and they will acquire burrows by grabbing, as shown in Eq.\ref{eq3}.
\begin{equation}
    X^{t+1}_{i,j}=X^{t}_{i,j}-X^t_{z,j}+X_{shade}
    \label{eq3}
\end{equation}
where $z$ is a randomly selected crayfish, calculated by Eq.\ref{eq4}.
\begin{equation}
    z = round(rand\times(N-1))+1
    \label{eq4}
\end{equation}

\begin{itemize}
    \item Foraging stage
\end{itemize}

When $temp\leq30$, it is suitable for crayfish to forage. They will move towards the food, and upon finding it, assess its size. The position of the food is defined in Eq.\ref{eq5}.
\begin{equation}
    X_{food} = X_G
    \label{eq5}
\end{equation}
Food size is defined as Eq.\ref{eq6}.
\begin{equation}
    Q=C_3\times rand\times (\frac{{fitness}_i}{{fitness}_{food}})
    \label{eq6}
\end{equation}
where $C_3$ is the food factor, which represents the largest food, setting the value to $3$; ${fitness}_i$ represents the fitness value of the $i$th crayfish; and ${fitness}_{food}$ represents the fitness value of the food.

When $Q>{\frac{C_3+1}{2}}$, it means that the food is relatively large, then the crayfish will tear the food, and the mathematical model of tearing is shown in Eq.\ref{eq7}.
\begin{equation}
    X_{food}=exp(-\frac{1}{Q})\times X_{food}
    \label{eq7}
\end{equation}
When the food is torn, the crayfish feeds alternately on the torn food, and the mathematical model of feeding is shown in Eq.\ref{eq8}.
\begin{equation}
    \begin{split}
        X^{t+1}_{i,j} &= X^t_{i,j} + X_{\text{food}} \cdot p \cdot \left( \cos\left( 2\pi \cdot \text{rand} \right) \right) \\
        &\quad - X_{\text{food}} \cdot p \cdot \left( \sin\left( 2\pi \cdot \text{rand} \right) \right)
    \end{split}
    \label{eq8}
\end{equation}
When $Q\leq{\frac{C_3+1}{2}}$, the crayfish only need to move toward the food and feed directly, and its mathematical model is shown in Eq.\ref{eq9}.
\begin{equation}
    X^{t+1}_{i,j} = (X^t_{i.j}-X_{food})\times p + p\times rand\times X^t_{i,j}
    \label{eq9}
\end{equation}

In the foraging phase, crayfish use different feeding methods according to the size of the food $Q$.
When the food is large, it is first torn into pieces and then fed; when the food is small, it is directly moved towards it to feed; through the foraging phase, the COA approximates the optimal solution to find the globally optimal solution.

\subsection{Optimizations on COA}

To enhance COA's performance in constraint and feature selection problems, many researchers have conducted optimization studies. Yi Zhang et al. \cite{zhang2024implementation} employed the Halton sequence to uniformly distribute the initial population in the search space and introduced the aggregation effect from the marine predator algorithm.
The resulting Enhanced COA (ECOA) showed a good balance between exploration and exploitation and achieved excellent results in the Three-Bar Truss Design and Pressure Vessel Design problems.

Heming Jia et al. further improved the COA with a novel environment - updating mechanism \cite{jia2024modified}.
They used the water quality factor to guide the search and integrated the ghost antagonism learning strategy.
This improved algorithm performed well on the CEC2020 test set and excelled in four constrained engineering and feature selection problems.

Lakhdar Chaib et al. \cite{chaib2024improved} combined fractional - order chaotic maps with COA to solve PV parameter estimation in single and dual - diode models.
They also used dimensional learning to enhance global and local search capabilities while maintaining population diversity.
Their algorithm outperformed others in accuracy, consistency, and convergence for PV parameter estimation.


In summary, while COA and its optimized variants have demonstrated advantages across diverse applications, critical challenges remain. 
Similar to many heuristic algorithms, the original COA depends on random population initialization, which can result in diversity loss. 
Additionally, there is a risk of local optima trapping due to an over-reliance on global-best solutions while fixed competition coefficients lead to slow convergence \cite{jafari2018cuckoo} \cite{li2024modified} \cite{wang2024improved}.  
MCOA addresses these challenges through multi-strategy enhancements, as detailed in Section \ref{sec:proposal}.
\section{MCOA}
\label{sec:proposal}
This paper proposes MCOA with three novel strategies aimed at addressing the key limitations of the original COA algorithm, enhancing its feasibility and effectiveness in numerical optimization and path planning. 
These strategies involve improvements in population initialization, exploration efficiency, and adaptive competition, with the goal of achieving faster convergence and higher solution quality.

\begin{figure}[hbtp]
    \centering
    \includegraphics[width=85mm]{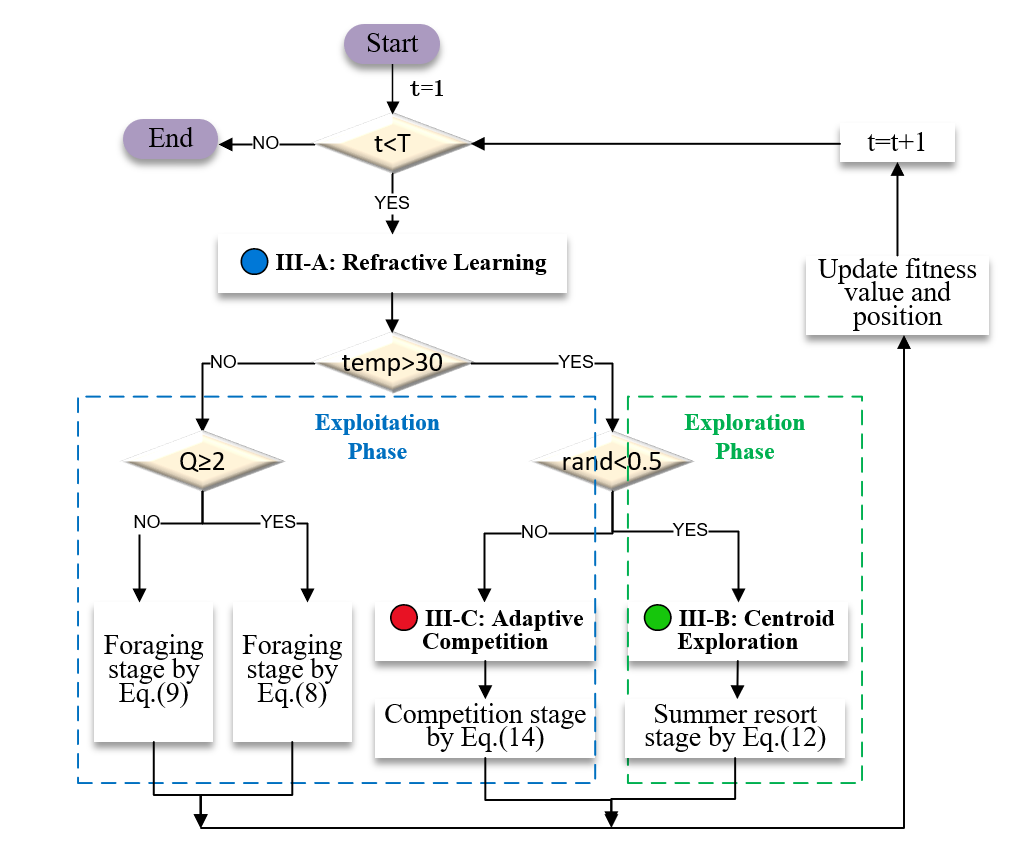}
    \caption{Flowchart of MCOA}
    \label{fig:3}
\end{figure}

As shown in Fig.\ref{fig:3}, first, Refraction Learning (\ref{sec:IIIA}) improves the initial population quality by dynamically generating reverse solutions Eq.\ref{eq10} and applying fitness-based selection Eq.\ref{eq11} to eliminate low-fitness individuals. 
The algorithm dynamically differentiates between stages using a temperature threshold (temp$>30$) and a random factor (rand$<0.5$). 
During the Exploration Phase(Summer Resort Stage), Centroid Exploration (\ref{sec:IIIB}), with its multi-path search Eq.\ref{eq12}, enhances global exploration capabilities, avoiding local optima.
In the Exploitation Phase, which is divided into the Competition Stage and the Foraging Stage, Adaptive Competition (\ref{sec:IIIC}) dynamically adjusts the search range using an adaptive competition mechanism Eq.\ref{eq14} and a time-decaying coefficient Eq.\ref{eq15}, gradually focusing on the refinement of the neighborhood around the current optimal solution. 
In the Foraging Stage, a dual-foraging strategy Eq.\ref{eq8} and \ref{eq9} further improve local search efficiency.
The details of each strategy are described below.

\subsection{Refractive opposition learning for population selection}
\label{sec:IIIA}
\begin{figure}[hbtp]
    \centering
    \includegraphics[width=80mm]{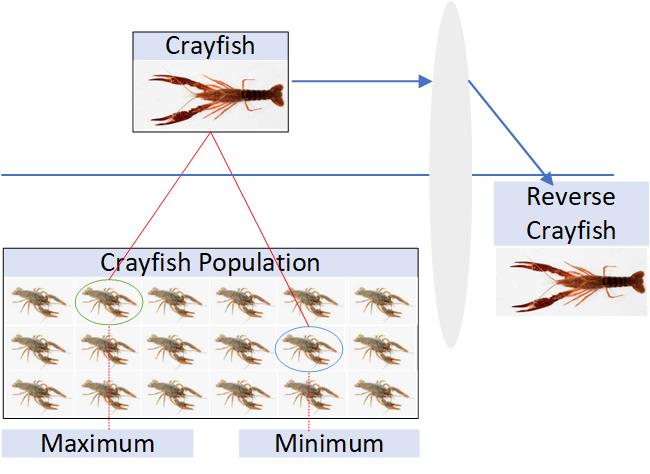}
    \caption{Refractive Learning for Population Selection}
    \label{fig:1}
\end{figure}

This subsection introduces a population selection strategy based on refractive opposition learning to address the issues of uneven quality and insufficient stability in the population initialization of the original COA algorithm. 
In the traditional COA algorithm, population initialization is achieved through randomization, where individuals obtain their starting positions randomly. 
Although this method is simple and convenient, it has significant drawbacks: the quality of randomly generated individuals can vary greatly, potentially leading to individuals with low fitness values in the initial population, which may negatively impact the algorithm's performance during early iterations.
Moreover, the randomness leads to different initial populations in each run, increasing uncertainty and instability in the results. 
To address these issues, the strategy incorporates a refractive learning mechanism during initialization to refine the population. 
Specifically, the refractive solutions of individuals are first calculated by Eq.\ref{eq10}, and then the better individual is selected between the refractive solution and the original solution. This approach effectively reduces the quality disparity among individuals in the initial population, maximizes the overall population quality, and thereby improves the algorithm's performance and stability.
\begin{equation}
    {reverse}_i=K\odot(MAX+MIN)-X_i
    \label{eq10}
\end{equation}
where ${reverse}_i$ denotes the oppositional solution obtained by particle $X_i$ after refractive oppositional-mutual learning; $K$ is a matrix of one row dim columns, where the elements in the matrix are random numbers between $0$ and $1$.
$MAX$ and $MIN$ are the maximum and minimum values of the individual, respectively. When the fitness value of the opposing solution is better than the original solution, the original solution is updated to the opposing solution; otherwise, no update is done. As shown in Eq.\ref{eq11}\cite{tu20243d}.
\begin{equation}
    X_i=
    \begin{cases}
      {reverse}_i,\ f(reverse(i))<f(X_i)\\
      X_i,\ other
    \end{cases} 
    \label{eq11}
\end{equation}
where $f(reverse(i))$ denotes the fitness value of the opposing solution and $f(X_i)$ denotes the fitness value of the original solution.
Fig.\ref{fig:1} graphically depicts this strategy for better understanding by the reader.

\subsection{Stochastic centroid-guided exploration}
\label{sec:IIIB}
\begin{figure}[hbtp]
    \centering
    \includegraphics[height=40mm]{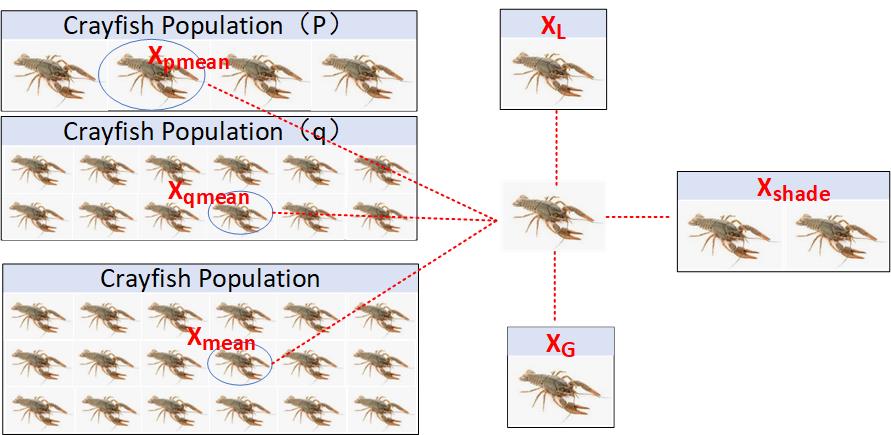}
    \caption{Six Random Crayfish Populations In Exploration}
    \label{fig:2}
\end{figure}

This subsection introduces a centroid-guided exploration based on stochastic factors to address the issue of the original COA algorithm falling into local optima during the exploration phase. 
In the exploration phase of the COA algorithm, the position update relies solely on the global optimal solution and the local optimal solution, which leads to overly concentrated search directions.
This may cause the population to prematurely converge to a local optimum while overlooking better global optimal solutions. 
Additionally, the lack of diversity in the search process further increases the risk of becoming trapped in local optima \cite{pierezan2018coyote}.
To address this issue, the proposed mechanism incorporates stochastic factors and multiple centroid information (e.g., \(X_{mean}\), \(X_{pmean}\), and \(X_{qmean}\)) as well as random positions (e.g., \(X_{shade}\)). By updating individual positions using Eq.\ref{eq12}, the mechanism enhances the population's exploration capability, reduces the risk of falling into local optima, and improves the algorithm's ability to discover better solutions.

\begin{equation}
    X^{t+1}_{i,j}=X^t_{i,j}+C_2\times rand\times(X_{{central(k1,j)}}-X^t_{i,j})
    \label{eq12}
\end{equation}
among them, $X_{(central(k1,j))}$ contains a total of $6$ location information, as shown in Eq.\ref{eq13}.
$k1$ is a random number between $1$ and $6$, which is used to decide which location in $X_{central}$ is selected for updating during the location update.
\begin{equation}
    X_{central}=\{X_G,X_L,X_{shade},X_{mean},X_{pmean},X_{qmean}\}
    \label{eq13}
\end{equation}
In this equation, $X_{mean}$ is the centroid of all individuals; $X_{pmean}$ is the centroid in population $p$; and $X_{qmean}$ is the centroid in population $q$.
Population $p$ is to randomly select $2\sim5$ individuals from the entire population, and $X_{pmean}$ is the centroid of this subset. Population $q$ is formed by selecting  $10\sim N$ individuals, and $X_{qmean}$ is the centroid of this larger subset.
$X_{shade}$ represents a random position that helps guide the algorithm out of local optima, adding diversity to the search and preventing premature convergence. 
Fig.\ref{fig:2} illustrates six types of random points controlled by the random factor, allowing the algorithm to escape local optima effectively.

\subsection{Adaptive competition-based selection}
\label{sec:IIIC}
This subsection proposes an adaptive competition coefficient based on the cave competition strategy to address the imbalance between exploration and exploitation in the original COA algorithm and accelerate convergence. 
In the exploitation phase, the original COA algorithm updates positions using $X_{shade}$, $X^t_{z,j}$, and random individuals. 
However, its fixed competition mechanism cannot dynamically adjust exploration and exploitation weights, causing over-reliance on local information and a tendency to fall into local optima  \cite{binev2004adaptive}. 
To address this, the new strategy introduces an adaptive competition coefficient CC, which dynamically adjusts competition intensity based on the current and maximum iteration numbers, as shown in Eq.\ref{eq14}. This mechanism promotes exploration early and exploitation later, guiding the algorithm to converge faster, avoid premature local optima, and improve solution quality.
\begin{equation}
    X^{t+1}_{i,j}=X_{shade}+CC*(X^t_{i,j}-X^t_{z,j})\odot randn
    \label{eq14}
\end{equation}
where $CC$ is the adaptive competition coefficient, which is calculated by Eq.\ref{eq15}.
$randn$ is a standard normal distribution to better guide individuals to update like near the optimal solution and further optimize the quality of the solution.
\begin{equation}
    CC={(1-\frac{t}{T})}^{\frac{2t}{T}}
    \label{eq15}
\end{equation}

\section{Evaluation}
\label{sec:evaluation}

In this section, MCOA algorithm is applied to 3D UAV and 2D mobile robot path planning to verify its practical effectiveness and answer two research questions (RQs) as follows:

\begin{itemize}
    \item RQ1: Is MCOA more efficient in terms of time compared to state-of-the-art algorithms?
    \item RQ2: Does MCOA generate higher quality paths compared to state-of-the-art algorithms? 
\end{itemize}

To systematically evaluate RQ1, the computational time of MCOA is measured and compared with other algorithms.
For RQ2, the paths generated by MCOA and other baseline algorithms are assessed based on path length cost, threat avoidance cost,
altitude constraints cost, flight angle cost, total flight cost, and obstacle avoidance smoothness. 
Two experiments are conducted: Experiment 1 examines both RQ1 and RQ2 in a 3D UAV path planning scenario\cite{zhao2018survey}. In order to further investigate the relationship between path quality and different map sizes, Experiment 2 is specifically designed to validate RQ2 in a 2D mobile robot path planning scenario\cite{pak2022field}.
Each algorithm is independently run $30$ times to ensure statistical reliability, with a population size of $50$ and $1000$ iterations per run.

\subsection{Experiment 1 - 3D UAV path planning}

\subsubsection{Exp1 - Settings}

MCOA is first evaluated in a 3D UAV path planning scenario, with detailed settings introduced as follows. The UAV flight environment is modeled through a composite function (Eq.\ref{eq16}-Eq.\ref{eq17}) considering realistic environmental obstacles, which subsequently derives four essential constraints(path length, threat avoidance, flight level, and flight angle) formulated in Eq.\ref{eq18} through Eq.\ref{eq28}. The algorithm is applied to solve the path planning problem within this constrained environment, where the flight path is defined between the starting point (150, 150, 50) and the endpoint (900, 720, 150).

\begin{itemize}
    \item UAV flight environment modeling
\end{itemize}

To simulate realistic flight conditions, the UAV flight environment is modeled using a composite function. The terrain and obstacles were defined as follows:
\begin{equation}
    \begin{split}
        z_1(x,y) &= 2(\cos x + \sin x) + \sin\left(\sqrt{x^2 + y^2}\right) \\
        &\quad + \cos\left(\sqrt{x^2 + y^2}\right)
    \end{split}
    \label{eq16}
\end{equation}
where $z_1(x,y)$ denotes the underlying terrain of the UAV flight environment.
Eq.\ref{eq17} is used to model obstacle factors such as hillsides:
\begin{equation}
    z_2(x,y)=\sum^{n'}_{i=1}h_iexp(-(\frac{x^2-x_i}{x_{si}})^2-(\frac{y^2-y_i}{y_{si}})^2)
    \label{eq17}
\end{equation}
where $n'$ is the number of slopes, $h_i$ is the height parameter of the slopes, $(x_i, y_i)$ are the coordinates of the center point of the slopes; and $x_{si}, y_{si}$ denote the amount of attenuation of the slopes in the direction of the x-axis and y-axis. $n'$ is set to $10$, $h_i$ is set as shown in Tab.\ref{table:1}.

\begin{table}[htbp]
\centering
    \caption{Peak Parameter Settings} 
    \begin{tabular}{ccccc}\\\hline 
        No. & x coordinate & y coordinate & z coordinate & radius\\\hline
        1 & 400 & 500 & 150 & 30\\
        2 & 700 & 150 & 150 & 50\\
        3 & 550 & 450 & 150 & 40\\
        4 & 350 & 100 & 150 & 50\\
        5 & 400 & 650 & 150 & 30\\
        6 & 800 & 800 & 150 & 30\\
        7 & 750 & 350 & 150 & 70\\
        8 & 150 & 350 & 150 & 60\\
        9 & 920 & 600 & 150 & 90\\
        10 & 920 & 200 & 150 & 50\\\hline
    \end{tabular}
    \label{table:1}
\end{table}

\begin{itemize}
    \item Modeling of UAV constraints
\end{itemize}

In addition to the primary map environment, natural environments typically include buildings, wires, and other obstacles. To simulate these conditions, the UAV flight constraints are modeled as follows.

Path length Constraint: Path distance is a critical factor in UAV path planning, directly impacting mission efficiency, safety, and energy consumption\cite{ghambari2024uav}. To minimize the total path length, the path shortest constraint is defined as shown in Eq.\ref{eq18}.
\begin{equation}
    F_1(X_i)=\sum^{n-1}_{j=1}L_{p_{i,j}p_{i,j+1}}
    \label{eq18}
\end{equation}
where $L_{p_{i,j}p_{i,j+1}}$ denotes the distance from node $i$ to node $i+1$, calculated by Eq.\ref{eq19}.
\begin{equation}
    \begin{split}
        L_{p_{i,j}p_{i,j+1}} &= \text{norm}\left( x_{i+1} - x_i; y_{i+1} - y_i; z_{i+1} - z_i \vphantom{\sqrt{x^2}} \right)
    \end{split}
    \label{eq19}
\end{equation}

Threat Avoidance Constraint: Avoiding threats such as buildings, trees, and power lines is crucial for UAV safety and mission success. The threat cost is calculated by \ref{eq20}.
\begin{equation}
    F_2(X_i)=\sum^{n-1}_{j=1}\sum^{K}_{k=1}T_k(p_{i,j}p_{i,j+1})
    \label{eq20}
\end{equation}
where $T_k(p_{i,j}p_{i,j+1})$ denotes flight constraint costs, calculated by Eq.\ref{eq21}. where $R_k$ is the radius of the $k$th cylindrical obstacle, $D$ is the collision region, and $d_k$ is the distance from the obstacle center to the path $L_{p_{i,j}p_{i,j+1}}$.
$Penalty$ is the threat cost penalty factor.
\begin{equation}
    T_k(p_{i,j}p_{i.j+1})=
    \begin{cases}
      0,\ d_k\geq D+R_k\\
      penalty*(D+R_k-d_k),\ R_k<d_k<D+R_k\\
      \infty,\ d_k\leq R_k
    \end{cases} 
    \label{eq21}
\end{equation}

Flight Level Constraints: In certain missions, UAVs must operate at specific altitudes. Adhering to airspace altitude restrictions is crucial for ensuring safe, efficient, and compliant UAV operations. The flight level constraint is represented by Eq.\ref{eq22}.
\begin{equation}
    F_3(X_i)=\sum^n_{j=1}H_{i,j}
    \label{eq22}
\end{equation}
where $H_{i,j}$ denotes the cost of the height of the $X_i$ location, calculated by Eq.\ref{eq23}. where $penalty$ is the penalty coefficient, $h_{ij}$ is the UAV's current altitude, $h_{min}$ is the minimum allowed altitude, and $h_{max}$ is the maximum allowed altitude.
\begin{equation}
    H_{i,j}=
    \begin{cases}
      penalty*(h_{i,j}-h_{max}),\ h_{i,j}\geq h_{max}\\
      0,\ h_{min}<h_{i,j}<h_{max}\\
      penalty*(h_{min}-h_{i,j}),\ 0<h_{i,j}\leq h_{min}
      \infty,\ h_{i,j}\leq 0
    \end{cases} 
    \label{eq23}
\end{equation}

Flight Angle Constraint: The UAV's horizontal turn angles $\alpha_{i,j}$ and vertical pitch angles $\beta_{i,j}$ are constrained to ensure smooth trajectories. The angle cost is calculated by Eq.\ref{eq24}. where  $a_1$ and $a_2$ denote the penalty coefficients for the UAV's horizontal turn angle and vertical pitch angle constraints respectively. where $\alpha_{i,j}$ and $\beta_{i,j}$ are calculated by Eq.\ref{eq25} and Eq.\ref{eq26}.
\begin{equation}
    F_4(X_i)=a_1\sum^{n-2}_{j=1}\alpha_{i,j}+a_2\sum^{n-1}_{j=1}|\beta_{i,j}-\beta_{i,j-1}|
    \label{eq24}
\end{equation}

\begin{equation}
  \alpha_{i,j}=arctan{\frac{L_{p'_{i,j}p'_{i,j+1}}\times L_{p'_{i,j}p'_{i,j+2}}}{L_{p'_{i,j}p'_{i,j+1}}\cdot L_{p'_{i,j}p'_{i,j+2}}}}
    \label{eq25}
\end{equation}

\begin{equation}
    \beta_{i,j}=arctan{\frac{Z_{i,j+1}-Z_{i,j}}{||L_{p_{i,j}p_{i,j+1}}||}}
    \label{eq26}
\end{equation}
where $L_{p'_{i,j}p'_{i.j+1}}$ is their projections on the plane, calculated by Eq.\ref{eq27}. where $k$ is the unit vector in the positive direction of the axis.
\begin{equation}
    L_{p'_{i,j}, p'_{i,j+1}} = k \times \left( L_{p_{i,j}, p_{i,j+1}} \times k \right)
    \label{eq27}
\end{equation}

Multi-factor flight cost function: Based on the four types of constraints, the total flight cost is calculated by Eq.\ref{eq28}.
\begin{equation}
    F(X_i)=\frac{\sum^4_{k=1}F_k(X_i)}{4}
    \label{eq28}
\end{equation}

\subsubsection{Exp1 - Results}
\begin{figure}[hbtp]
    \centering
    \includegraphics[height=45mm]{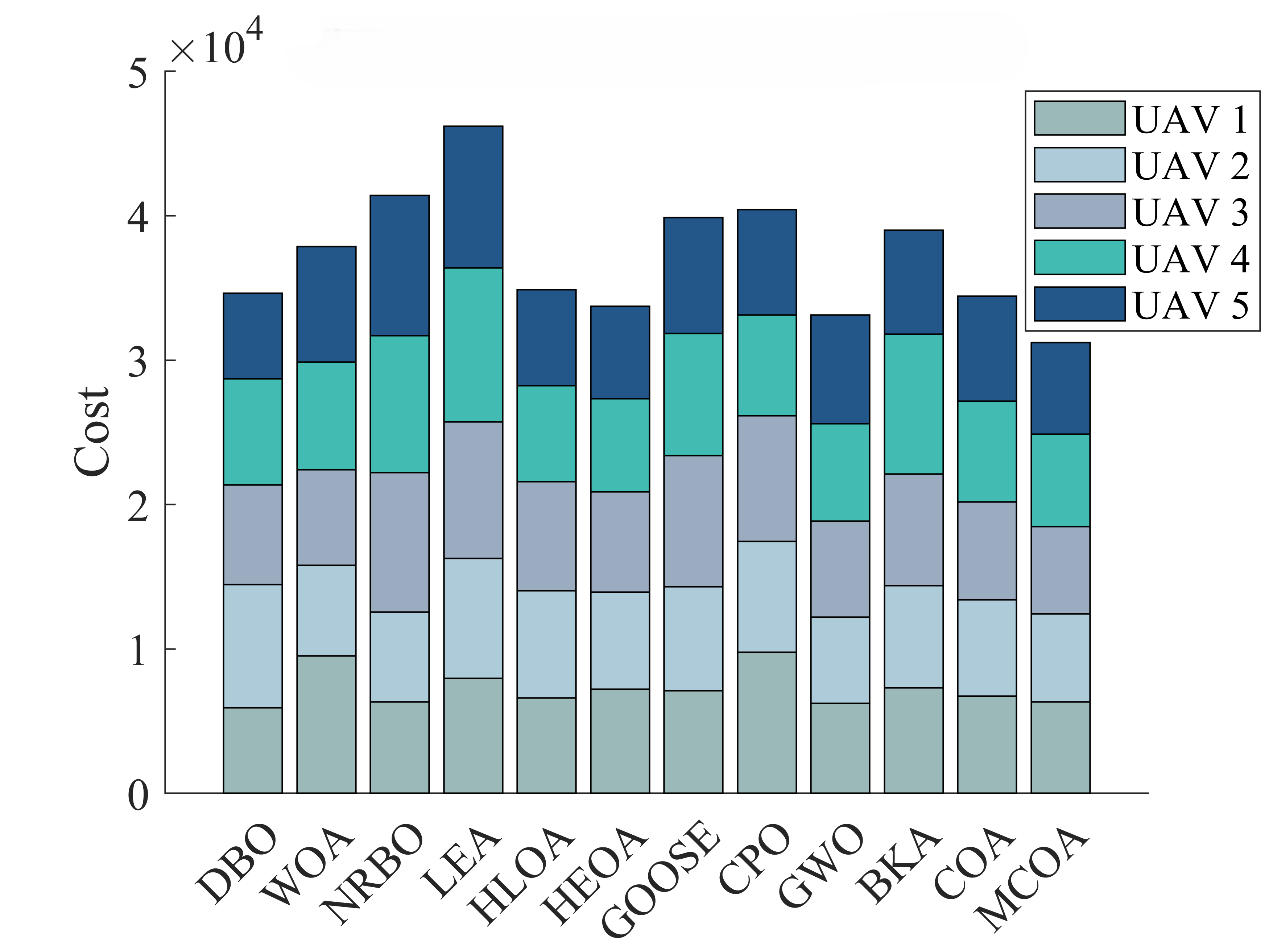}
    \caption{Comparison in Total Flight Cost}
    \label{fig:4}
\end{figure}

\begin{figure}[htbp]
    \centering
    \begin{subfigure}[t]{0.32\textwidth}
        \includegraphics[width=\linewidth,height=45.5mm]{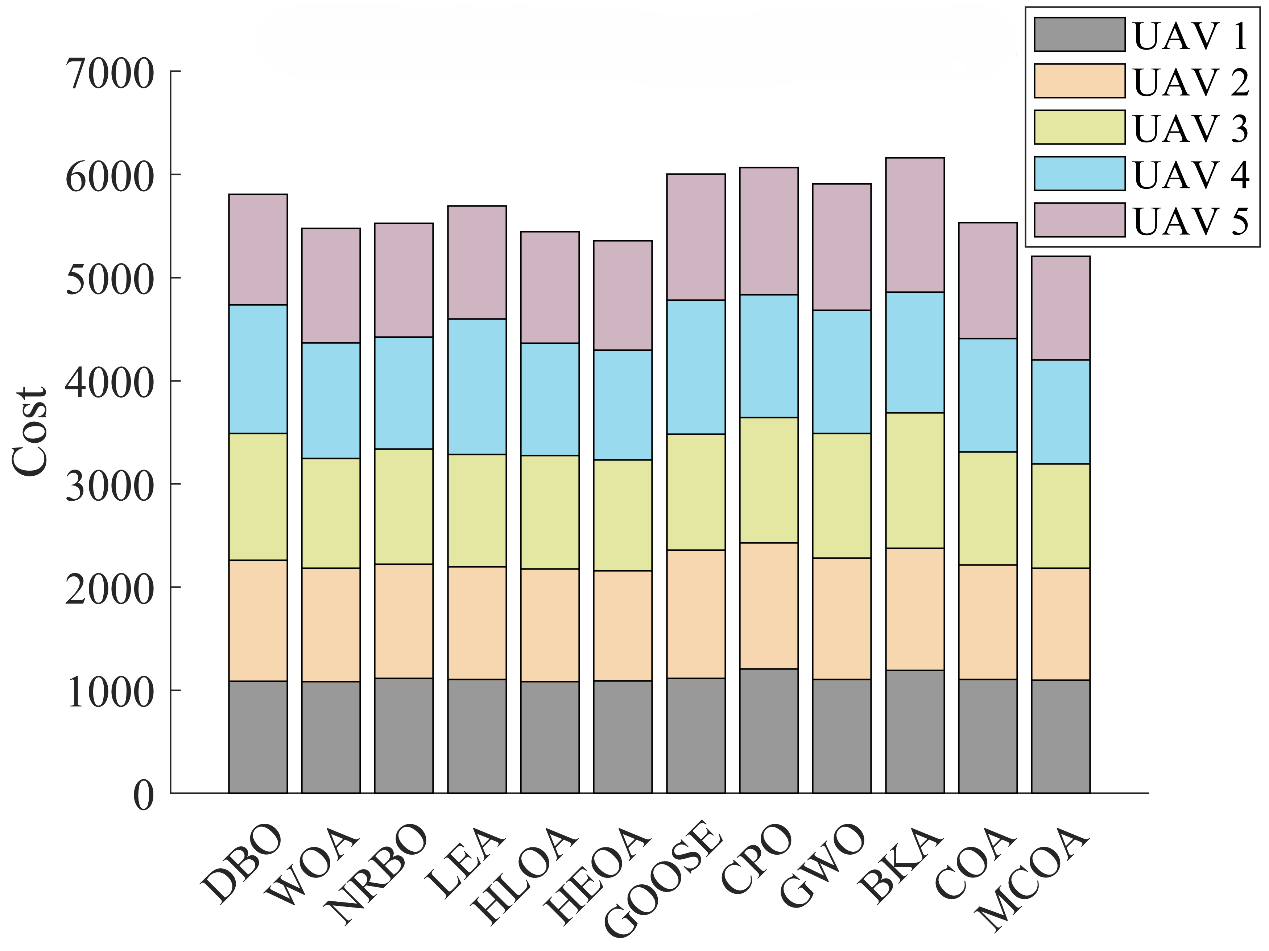}
        \caption{Path Length Cost}
        \label{fig:5}
    \end{subfigure}
    
    \begin{subfigure}[t]{0.32\textwidth}
        \includegraphics[width=\linewidth,height=46mm]{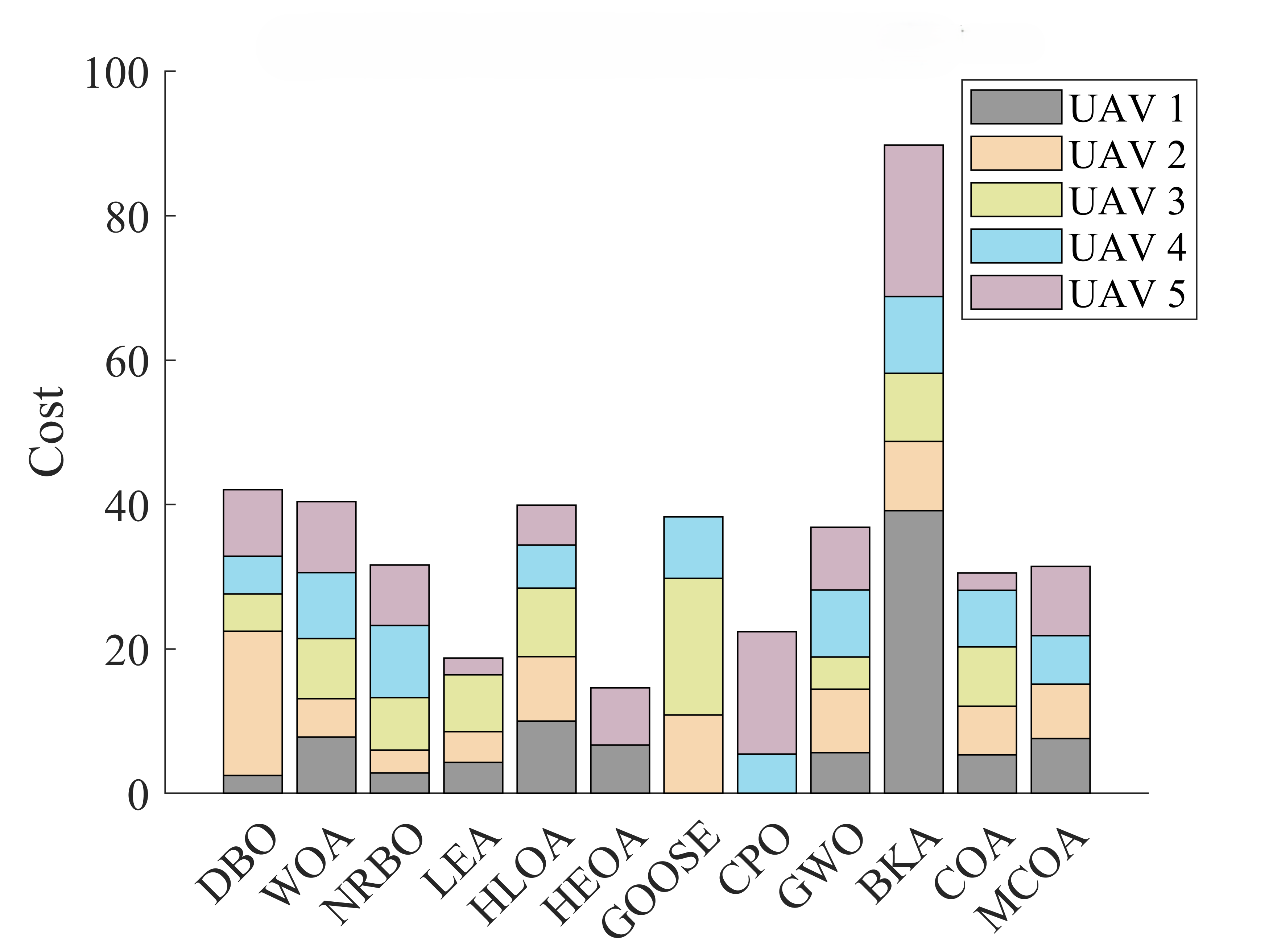}
        \caption{Threat Avoidance Cost}
        \label{fig:6}
    \end{subfigure}
    \begin{subfigure}[t]{0.32\textwidth}
        \includegraphics[width=\linewidth,height=46mm]{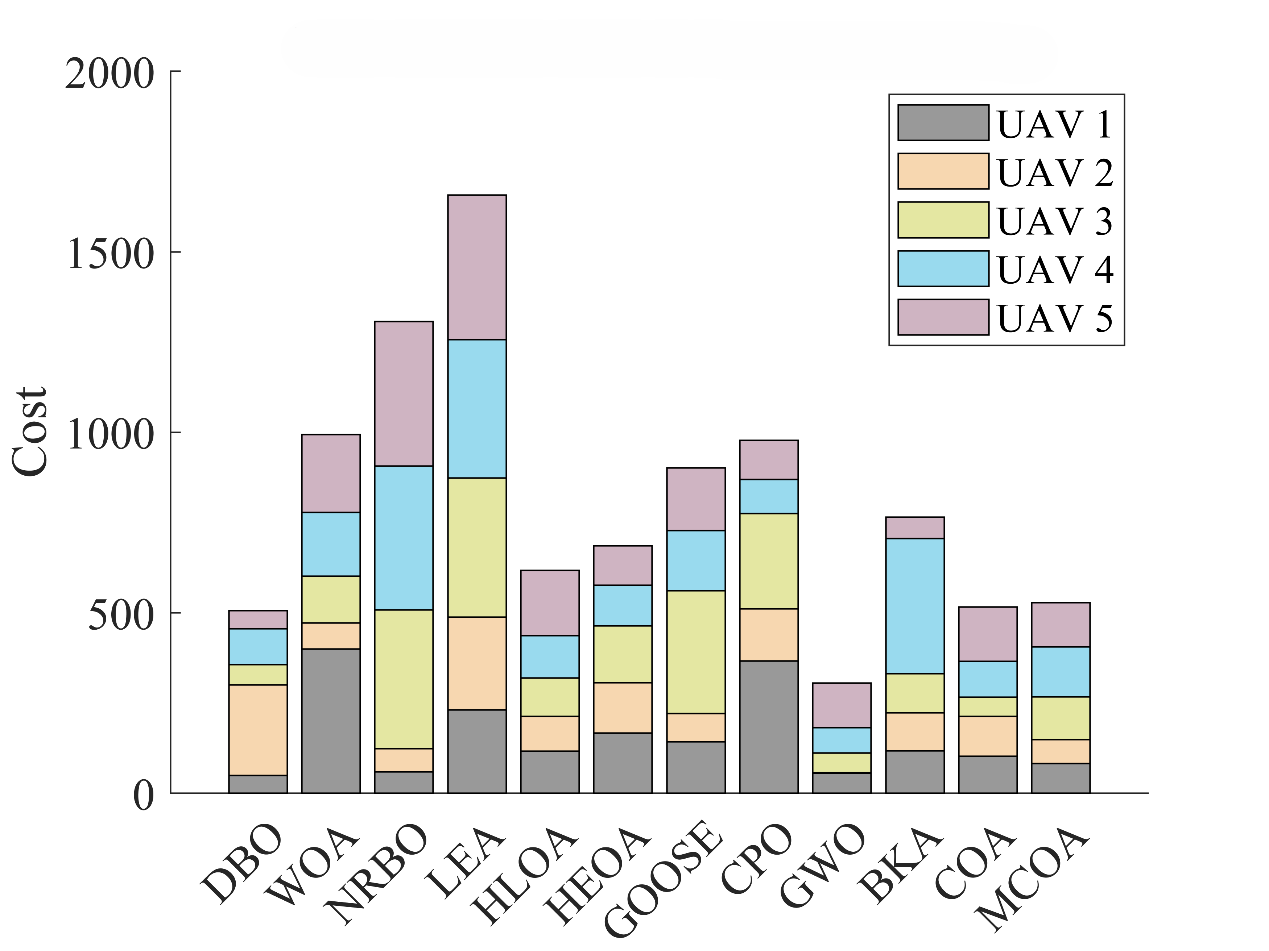}
        \caption{Flight Level Cost}
        \label{fig:7}
    \end{subfigure}
    
    \begin{subfigure}[t]{0.32\textwidth}
        \includegraphics[width=\linewidth,height=45.5mm]{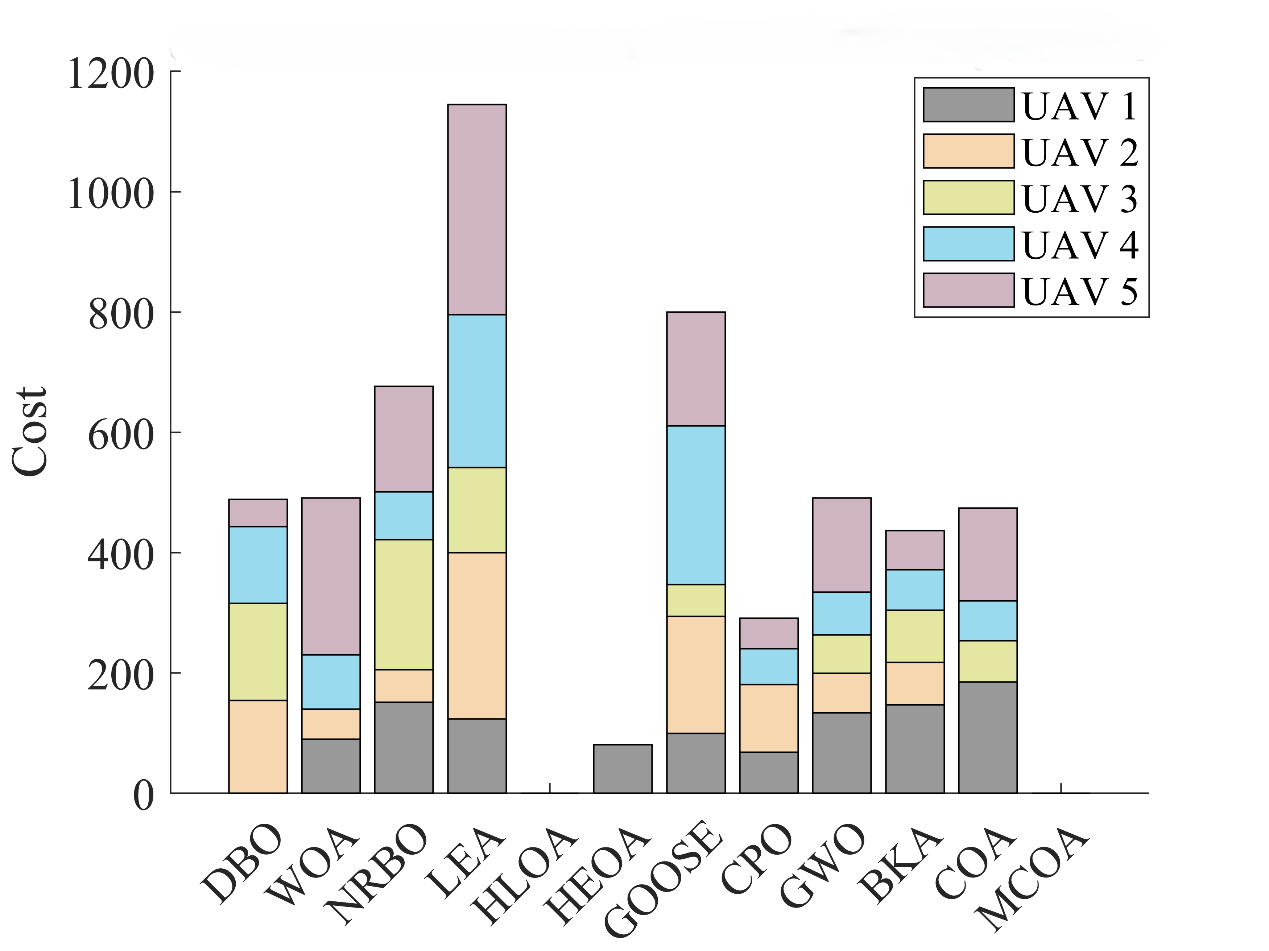}
        \caption{Flight Angle Cost}
        \label{fig:8}
    \end{subfigure}

    \caption{Comparison in Four Types of Costs}
    \label{fig:5-8}
\end{figure}

\begin{figure*}[htbp]
    \centering
    
    \begin{subfigure}[t]{0.32\textwidth}
        \includegraphics[width=\linewidth]{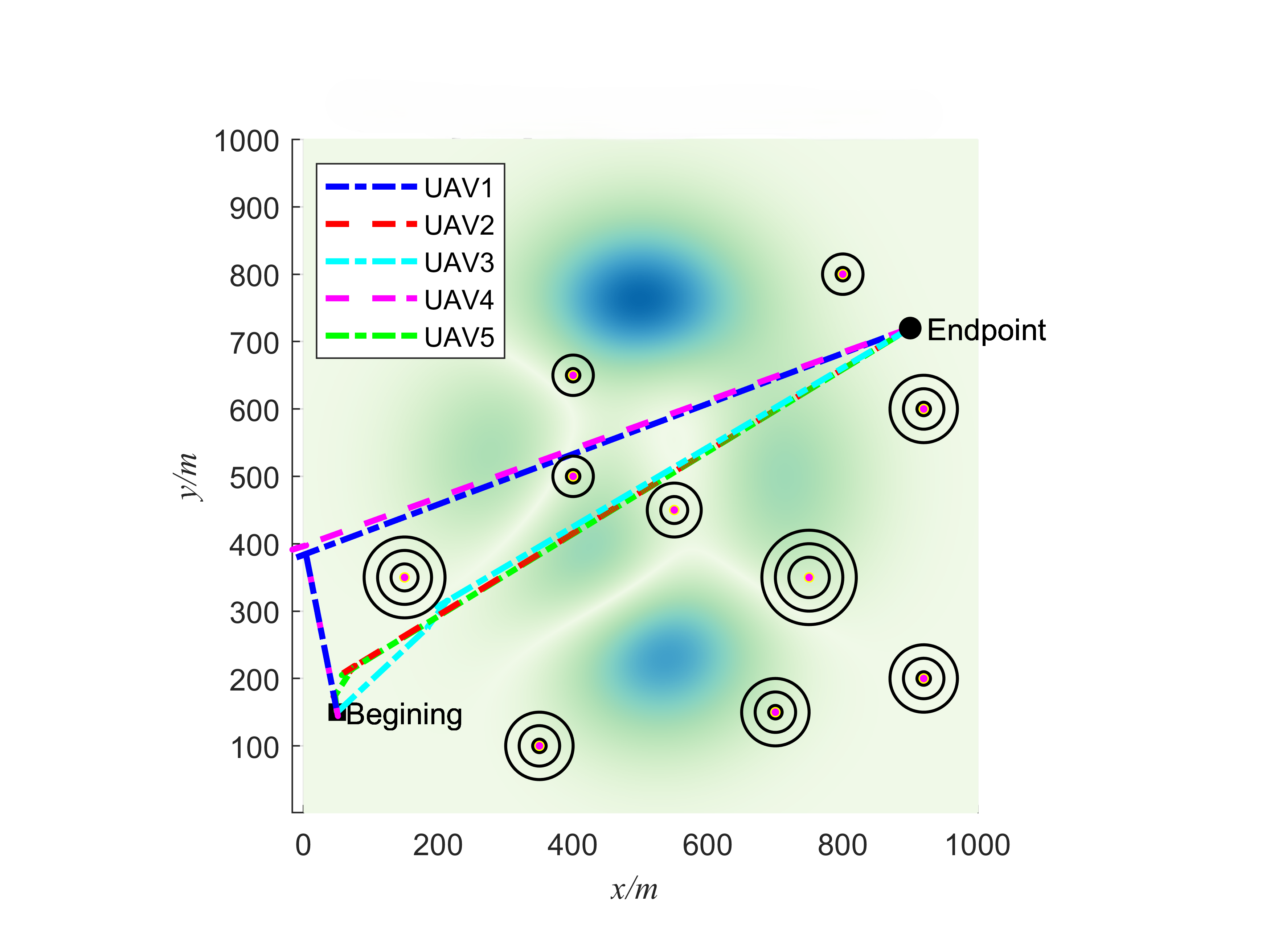}
        \caption{DBO}
        \label{fig:9a}
    \end{subfigure}
    \begin{subfigure}[t]{0.32\textwidth}
        \includegraphics[width=\linewidth]{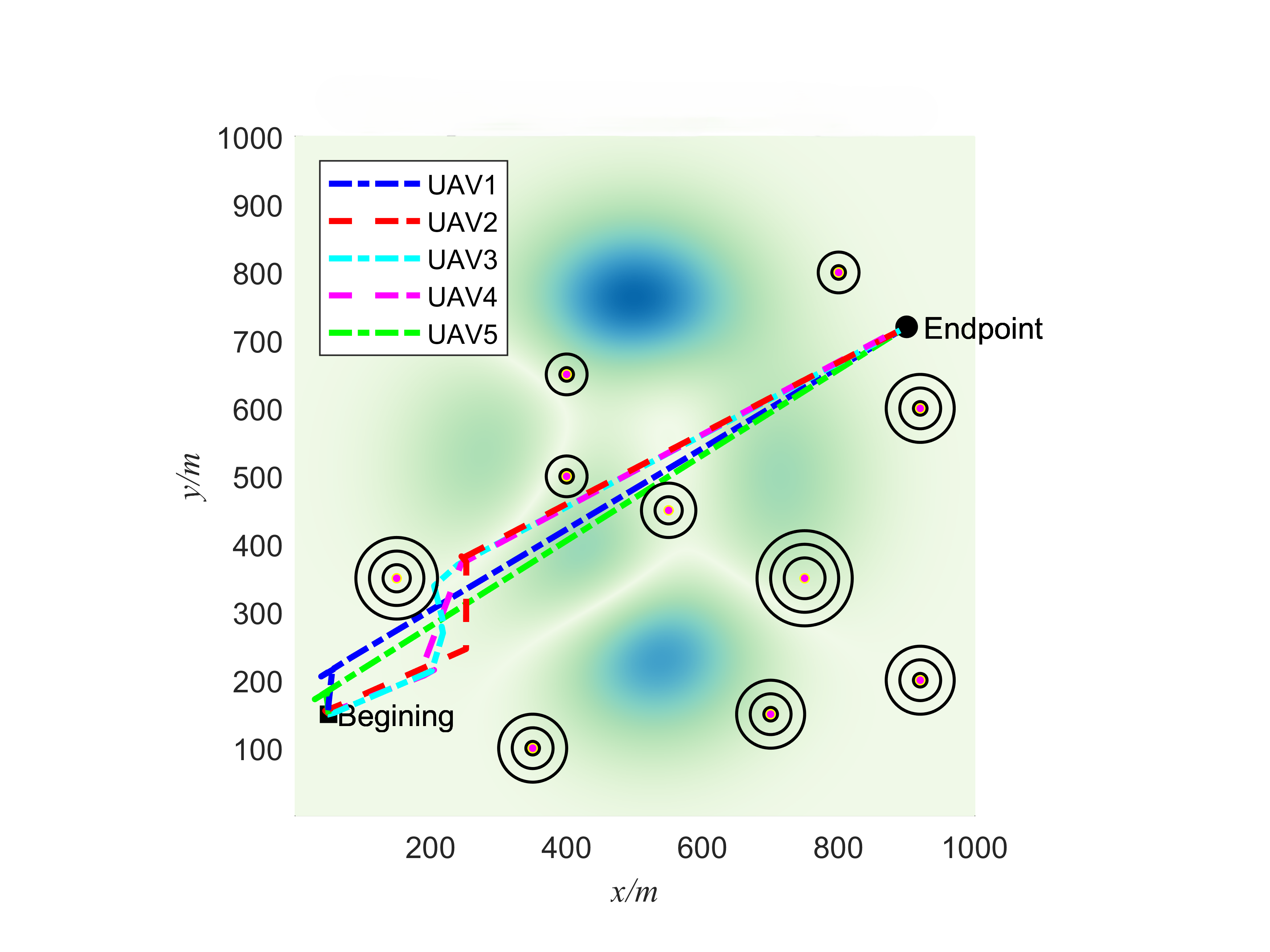}
        \caption{WOA}
        \label{fig:9b}
    \end{subfigure}
    \begin{subfigure}[t]{0.32\textwidth}
        \includegraphics[width=\linewidth]{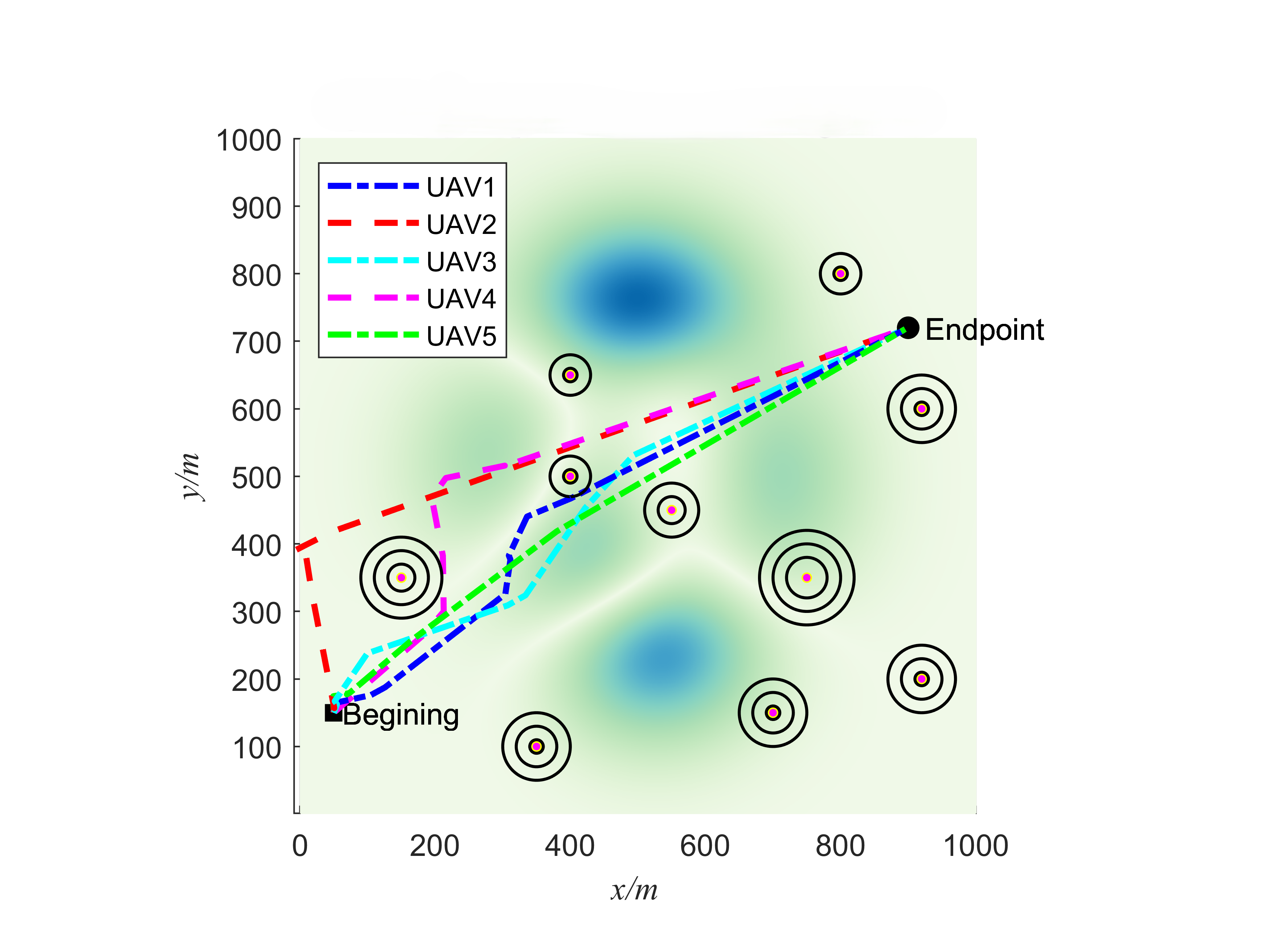}
        \caption{NRBO}
        \label{fig:9c}
    \end{subfigure}
    
    \begin{subfigure}[t]{0.32\textwidth}
        \includegraphics[width=\linewidth]{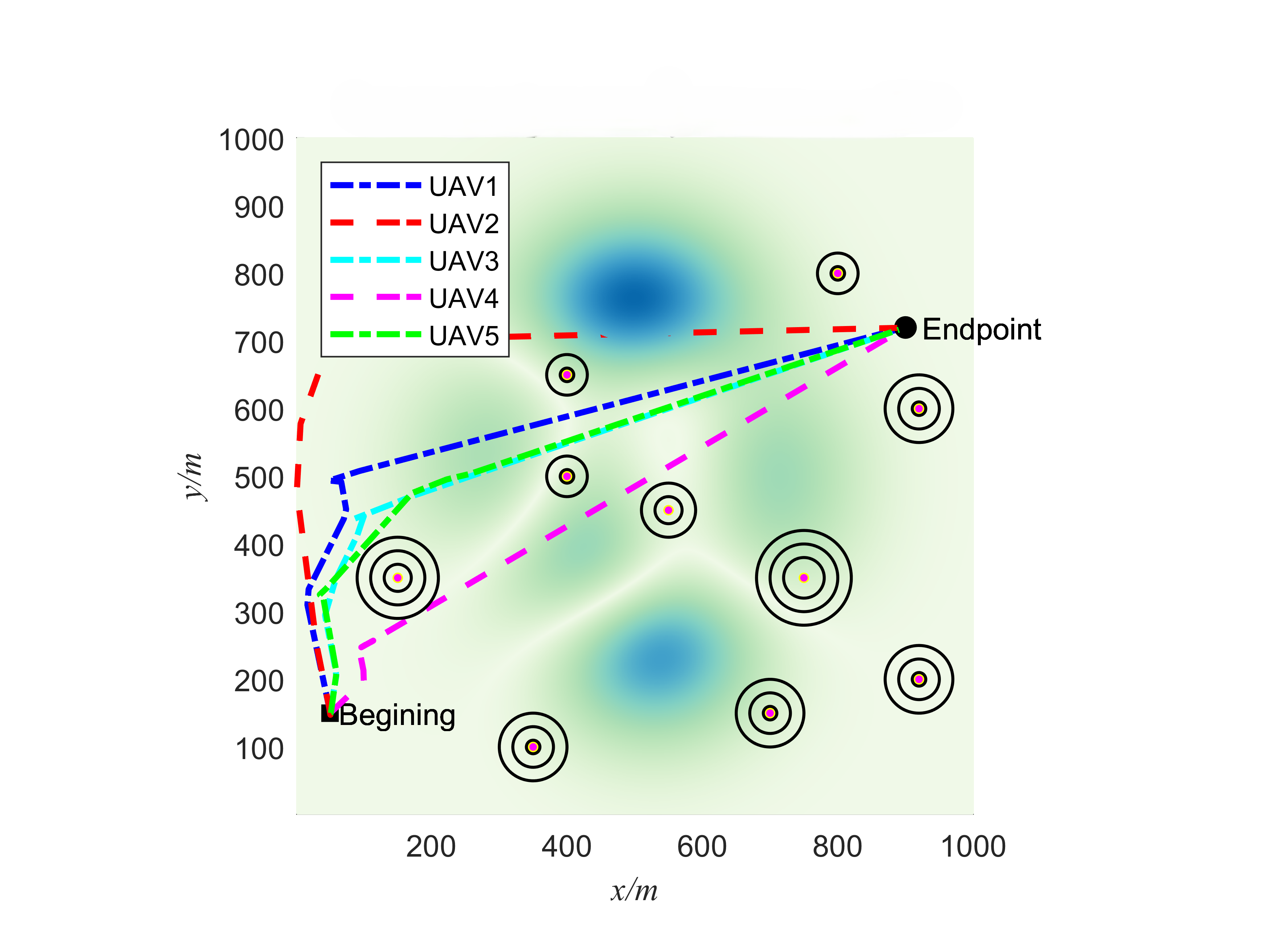}
        \caption{LEA}
        \label{fig:9d}
    \end{subfigure}
    \begin{subfigure}[t]{0.32\textwidth}
        \includegraphics[width=\linewidth]{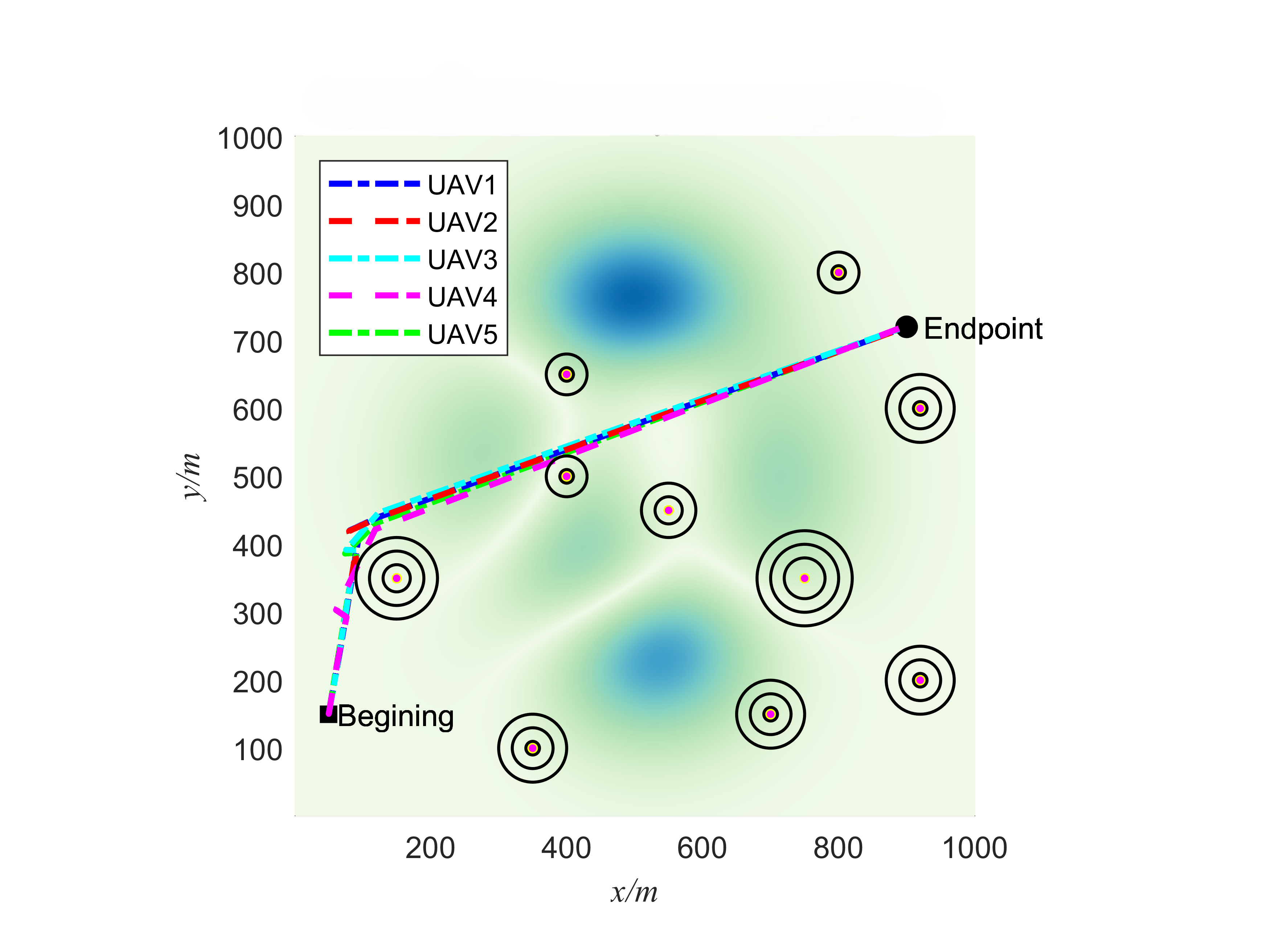}
        \caption{HLOA}
        \label{fig:9e}
    \end{subfigure}
    \begin{subfigure}[t]{0.32\textwidth}
        \includegraphics[width=\linewidth]{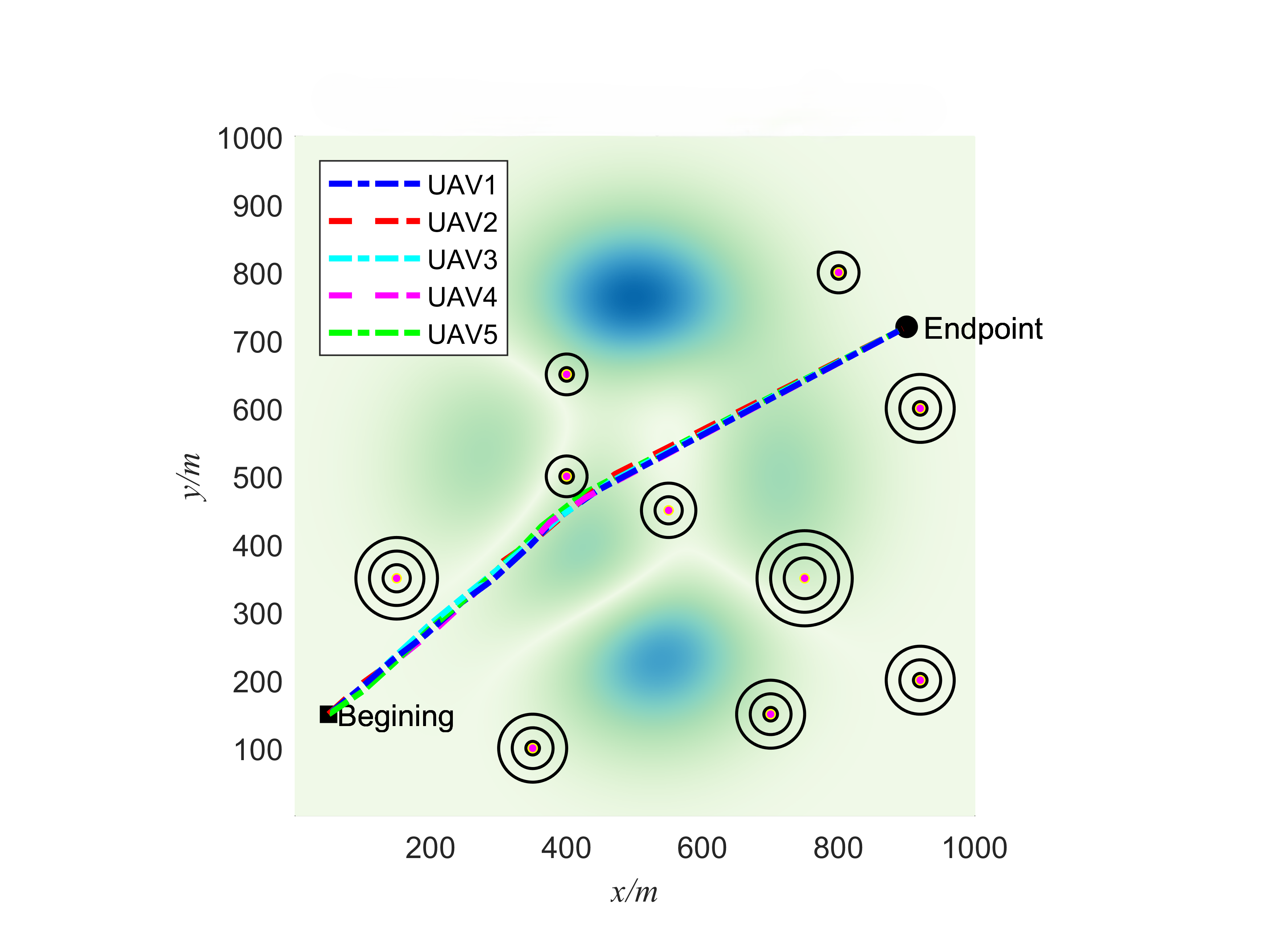}
        \caption{HEOA}
        \label{fig:9f}
    \end{subfigure}
    
    \begin{subfigure}[t]{0.32\textwidth}
        \includegraphics[width=\linewidth]{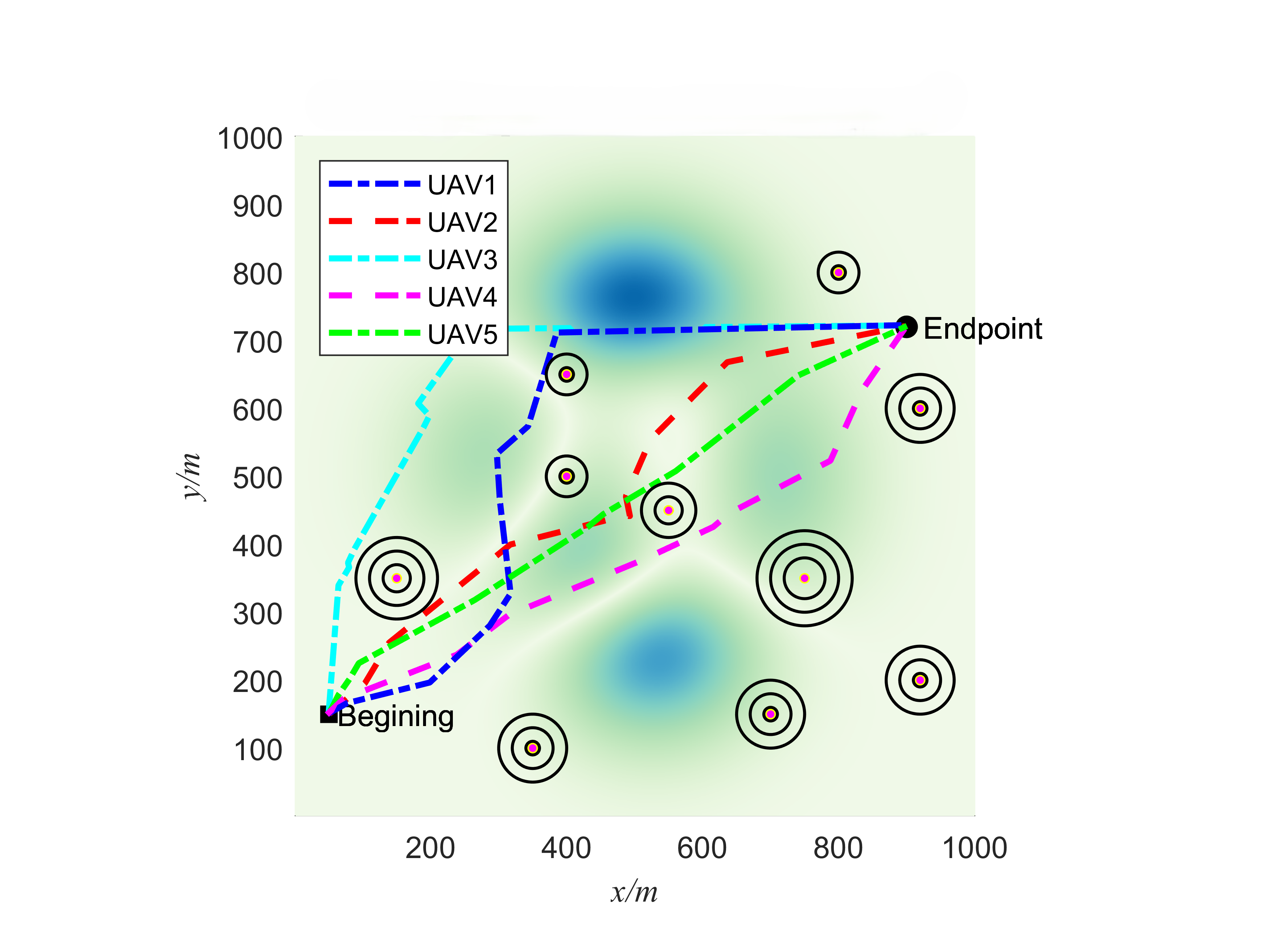}
        \caption{GOOSE}
        \label{fig:9g}
    \end{subfigure}
    \begin{subfigure}[t]{0.32\textwidth}
        \includegraphics[width=\linewidth]{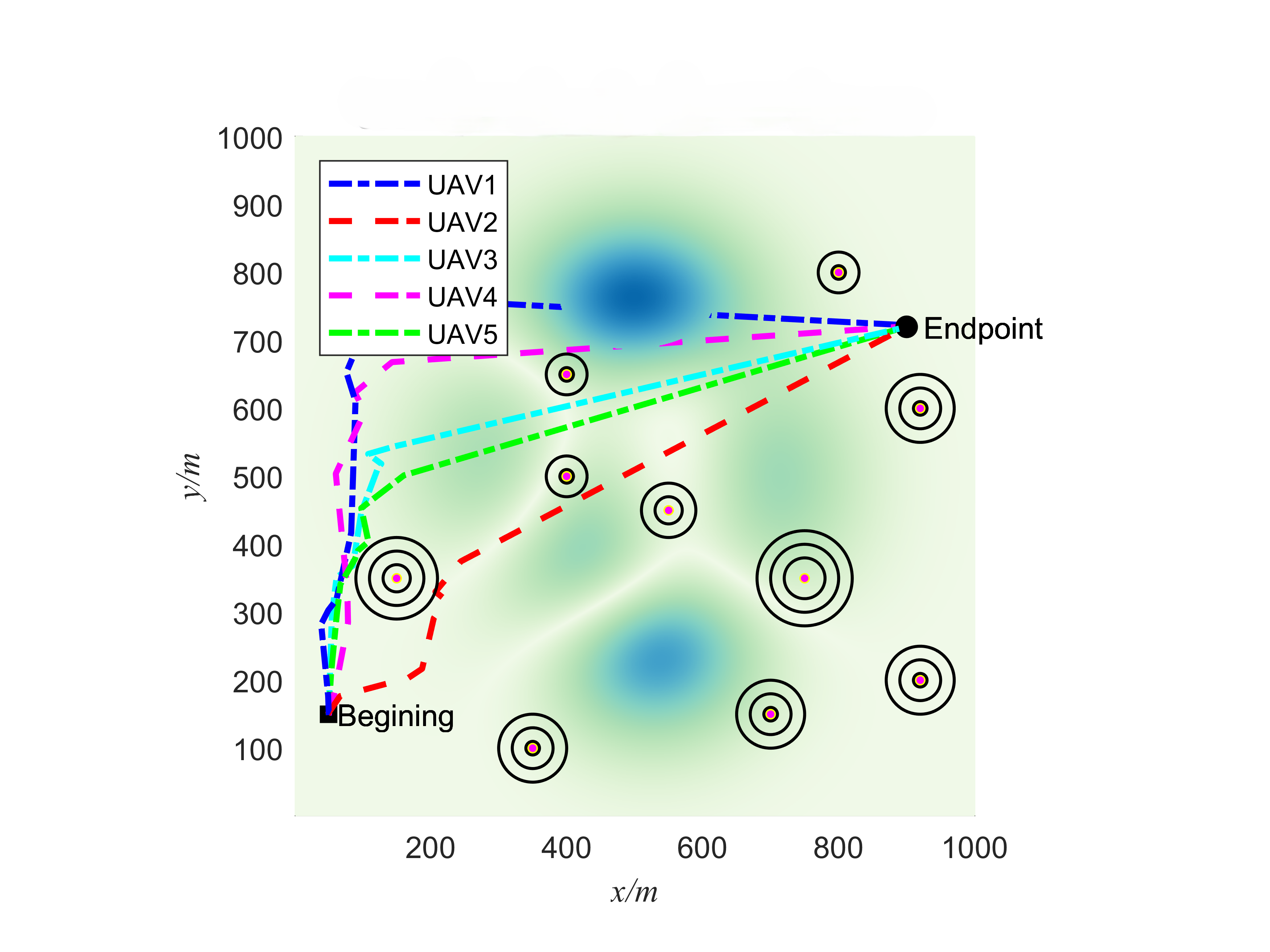}
        \caption{CPO}
        \label{fig:9h}
    \end{subfigure}
    \begin{subfigure}[t]{0.32\textwidth}
        \includegraphics[width=\linewidth]{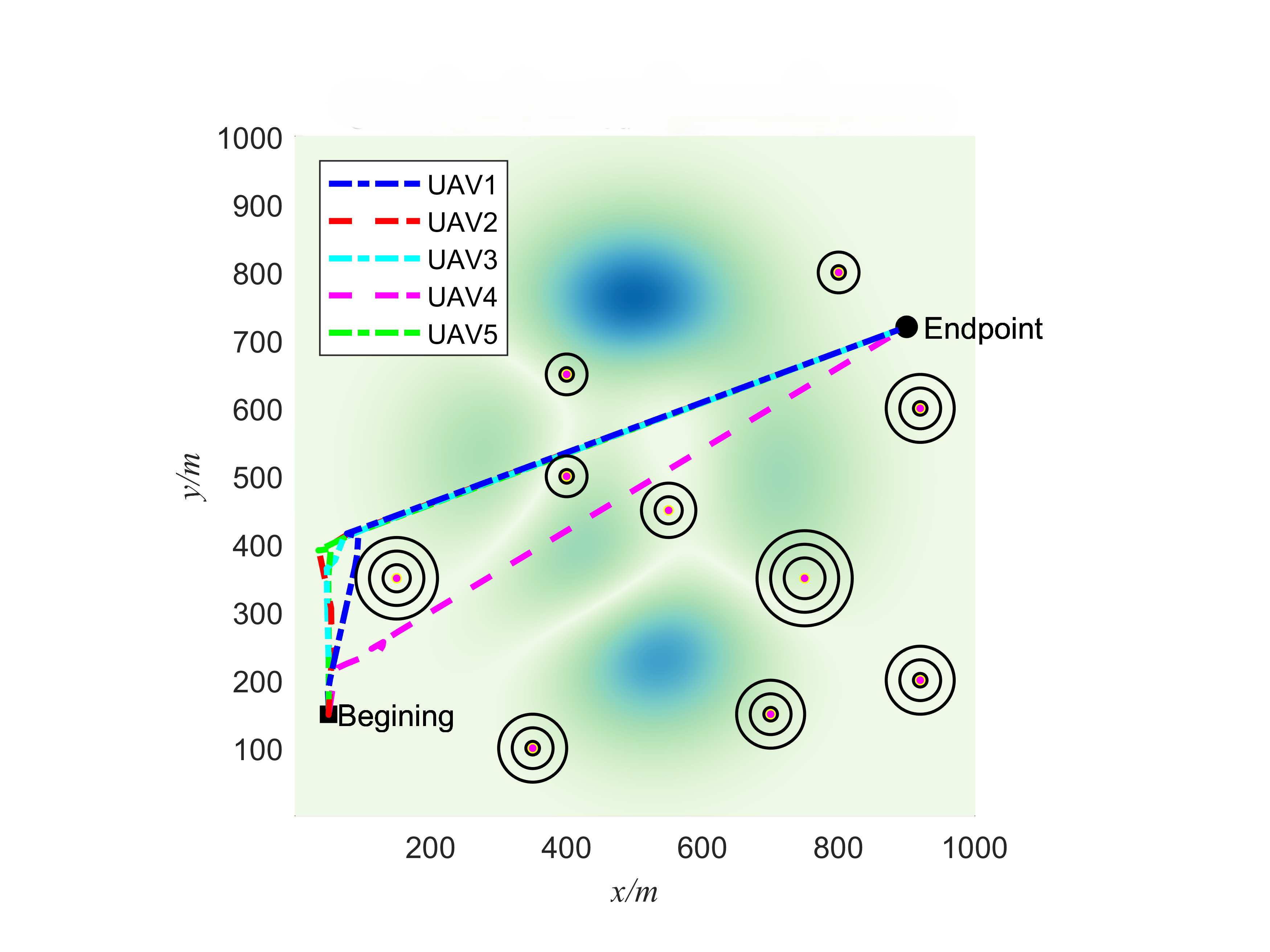}
        \caption{GWO}
        \label{fig:9i}
    \end{subfigure}
    
    \begin{subfigure}[t]{0.32\textwidth}
        \includegraphics[width=\linewidth]{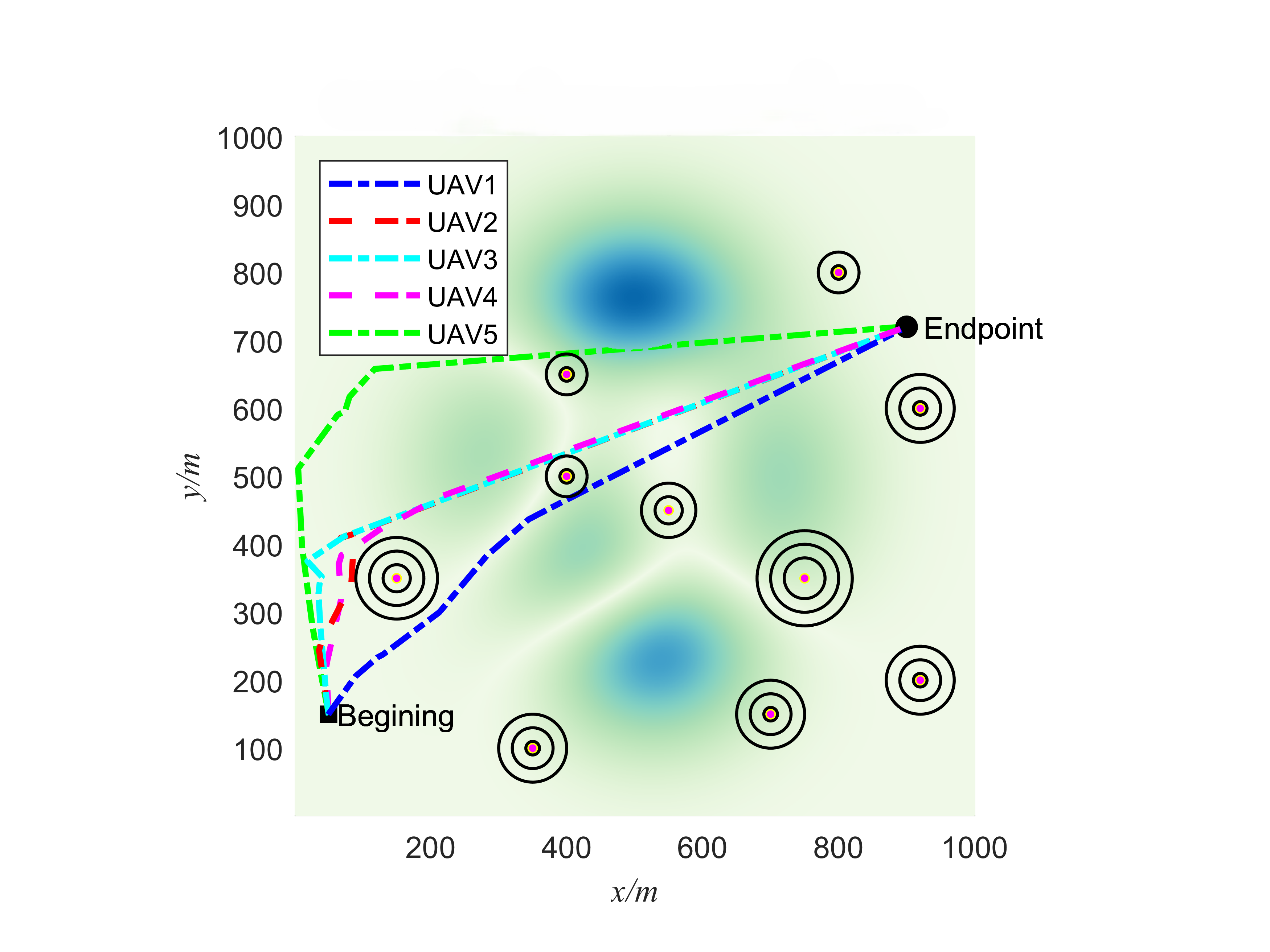}
        \caption{BKA}
        \label{fig:9j}
    \end{subfigure}
    \begin{subfigure}[t]{0.32\textwidth}
        \includegraphics[width=\linewidth]{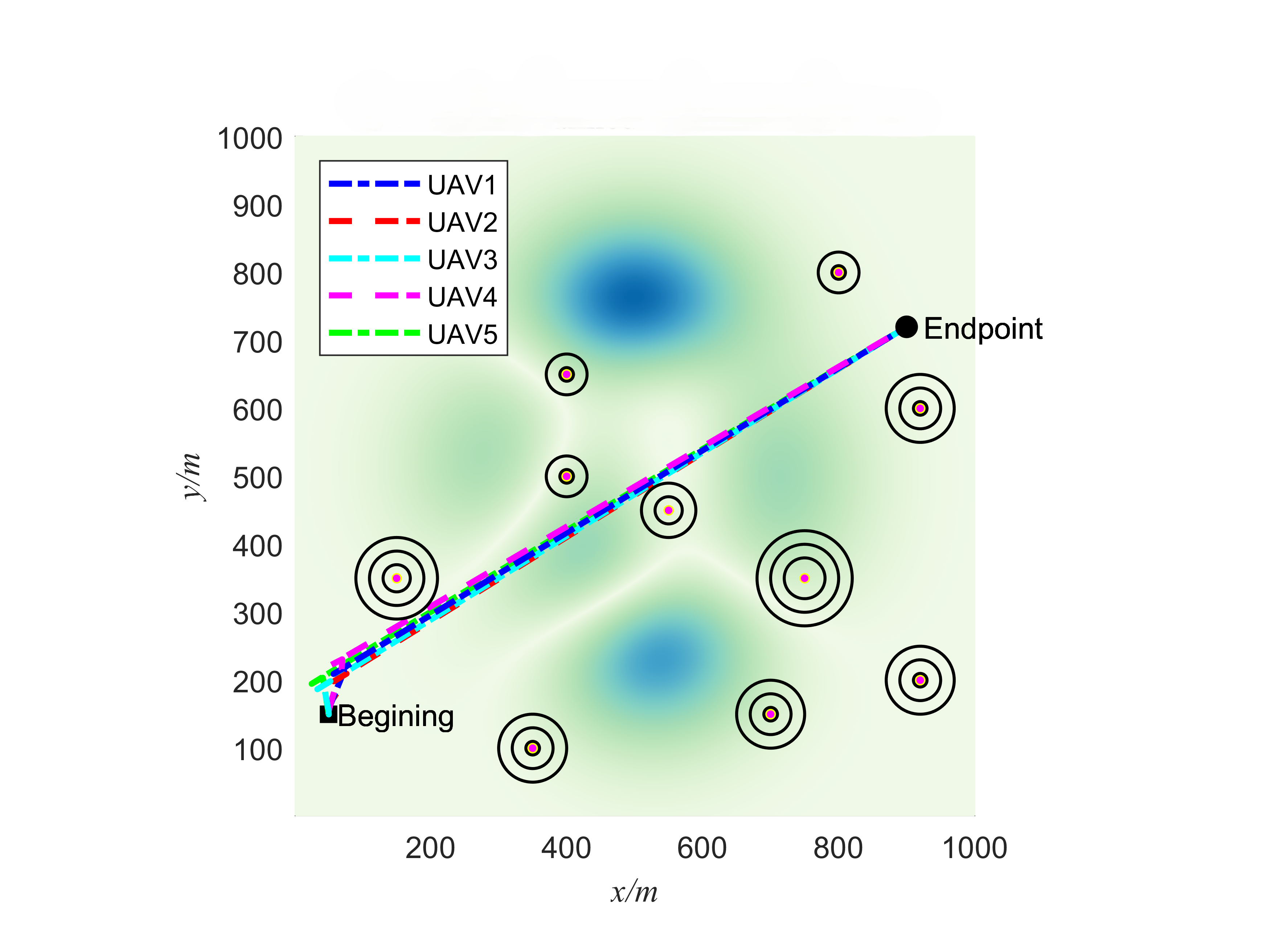}
        \caption{COA}
        \label{fig:9k}
    \end{subfigure}
    \begin{subfigure}[t]{0.32\textwidth}
        \includegraphics[width=\linewidth]{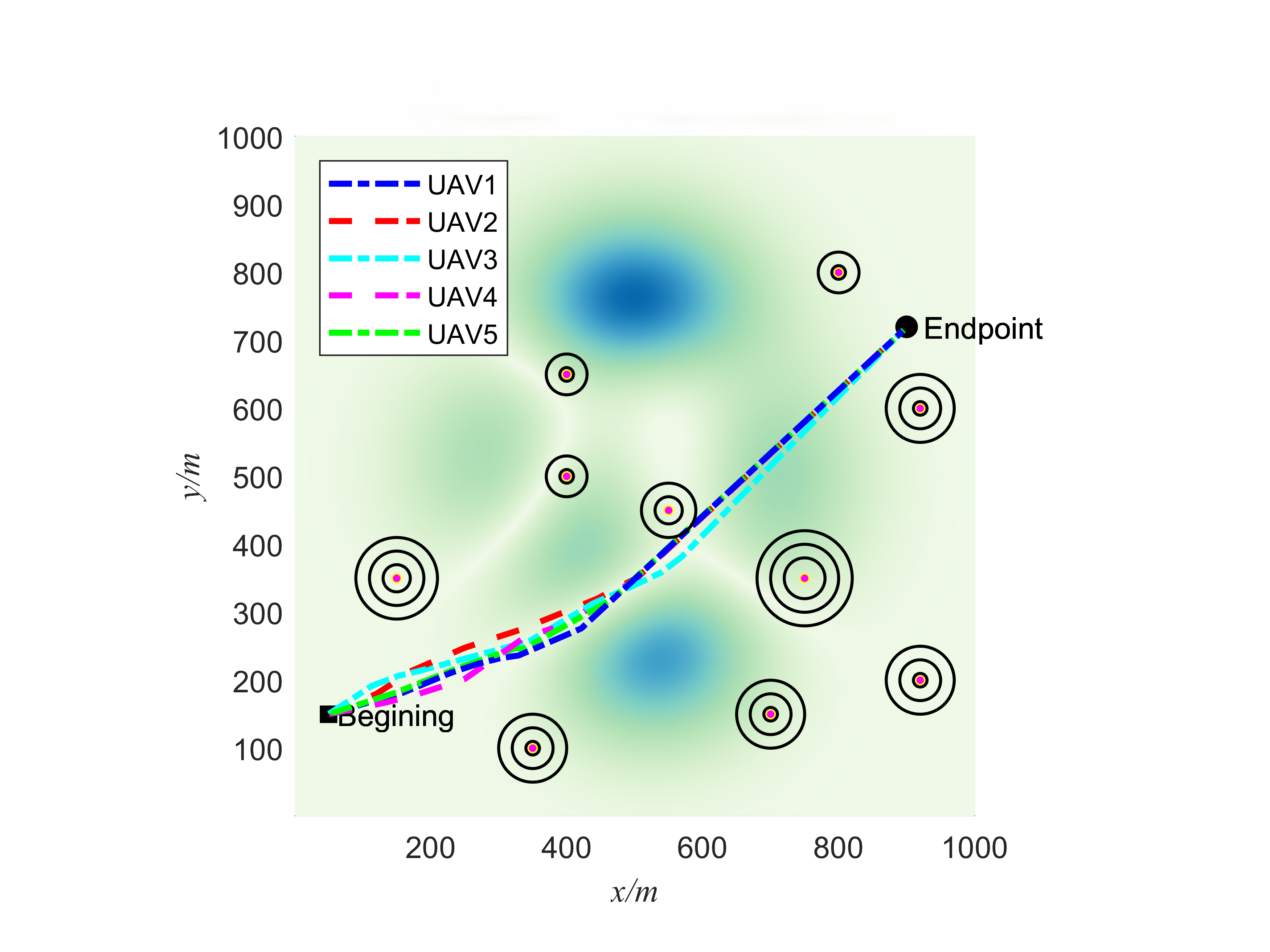}
        \caption{MCOA}
        \label{fig:9l}
    \end{subfigure}
    
    \caption{UAV Trajectories of Different Algorithms}
    \label{fig:9}
\end{figure*}

Fig.\ref{fig:4} shows the total cost of each algorithm for path planning for $5$ UAVs, with MCOA achieving the lowest overall cost, which is 11\% lower than the second-best algorithm. 
Fig.\ref{fig:5-8} shows the cost of each of the five UAVs in terms of path length, threat avoidance, flight level, and flight angle, respectively.
MCOA demonstrates advantages across all cost metrics. 
Specifically, the total cost of MCOA is 16.7\% lower than the average and 8.8\% lower than the second-best algorithm. In terms of path cost, MCOA achieves the lowest value, which is 9.1\% lower than the average. For angle cost, MCOA's cost is $0$, significantly lower than the average cost of other algorithms. Fig.\ref{fig:9}  shows the path trajectories planned by each algorithm. MCOA generated smoother paths than the other competing algorithms, highlighting its ability to produce high-quality paths in UAV path planning.
In terms of time consumption, as shown in Tab.\ref{table_T}, MCOA took $13.80$ s, making it the fastest among all the algorithms. The average time consumption of these algorithms is approximately $44.91$ s, with MCOA demonstrating a 69.2\% improvement in computational efficiency compared to the average.

\begin{table*}[htbp]
\centering
    \caption{UAV Path Planning Time Consumption} 
    \begin{tabular}{ccccccccccccc}\\\hline 
        Metric/Algorithm & DBO & WOA & NRBO & LEA & HLOA & HEOA & GOOSE & CPO & GWO & BKA & COA & MCOA\\\hline
        Time consumption (s) & 14.33 & 13.84 & 15.51 & 16.32 & 17.59 & 181.14 & 44.69 & 42.19 & 84.51 & 78.18 & 15.85 & \textbf{13.80}\\
    \hline
    \end{tabular}
    \label{table_T}
\end{table*}

\subsubsection{Exp1 - Discussions}
In the 3D UAV path planning experiments, MCOA significantly reduced path costs and generated high-quality paths while satisfying multiple constraints in complex environments. Experimental data showed that the total path cost of MCOA is 16.7\% lower than the average, effectively reducing flight distance and energy consumption while shortening mission execution time, making it particularly suitable for time-sensitive tasks. Furthermore, the flight trajectories planned by MCOA are smoother, with minimal angle fluctuations, significantly improving stability and reducing mechanical wear. These advantages make MCOA particularly well-suited for high-mobility mission scenarios, such as low-altitude reconnaissance and precision delivery in complex environments.
 
However, experimental results indicate that in complex environments, MCOA's improvement in controlling threat cost and high cost is limited compared to COA. Particularly in areas with dense obstacles, MCOA's ability to handle dynamic threats is slightly insufficient, and its flexibility in altitude adjustment requires further enhancement. For instance, when encountering suddenly elevated obstacles, MCOA's UAV altitude adjustment delay is, on average, $0.5$ seconds longer than that of COA, which may increase the difficulty of obstacle avoidance in high-speed flight scenarios. By further optimizing threat cost and altitude cost, MCOA is expected to achieve better performance in complex and dynamic environments.

\subsection{Experiment 2 - 2D Mobile robot path planning}

\subsubsection{Exp2 - Settings}
In this subsection, the MCOA algorithm is applied to 2D mobile robot path planning, modeling the robot's motion environment using a raster map method. The environment is divided into fixed-size grids, where each grid contains region information and uses $0$ and $1$ to represent its state: $1$ indicates an obstacle, and $0$ indicates a passable area. The robot movement path length is calculated using Eq.\ref{eq29}.
MCOA algorithm is evaluated against 11 other algorithms in three grid environments of varying sizes: $20*20$, $40*40$, and $60*60$. For each environment, the start point is set to ($1$,$1$) and the endpoint to $(n,n)$, where $n$ is the side length of the grid.
\begin{equation}
    L=\sum^n_{i=1}\sqrt{(x_{i+1}-x_i)^2-(y_{i+1}-y_i)^2}
    \label{eq29}
\end{equation}

\subsubsection{Exp2 - Results}
\begin{figure}[hbtp]
    \centering
    \includegraphics[height=45mm]{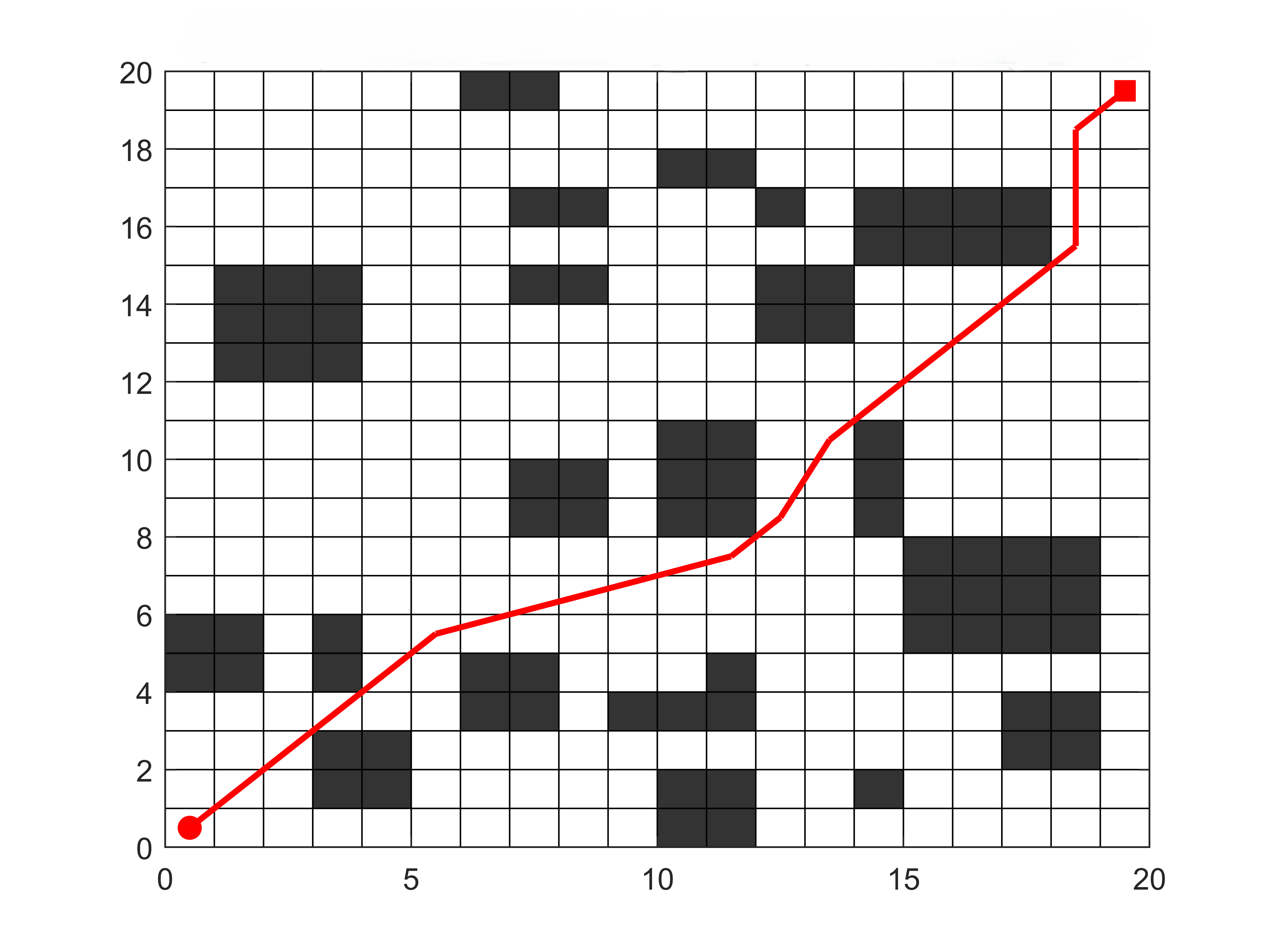}
    \caption{Robot Motion Trajectory In $20*20$ Grid}
    \label{fig:10}
\end{figure}

\begin{figure}[hbtp]
    \centering
    \includegraphics[height=45mm]{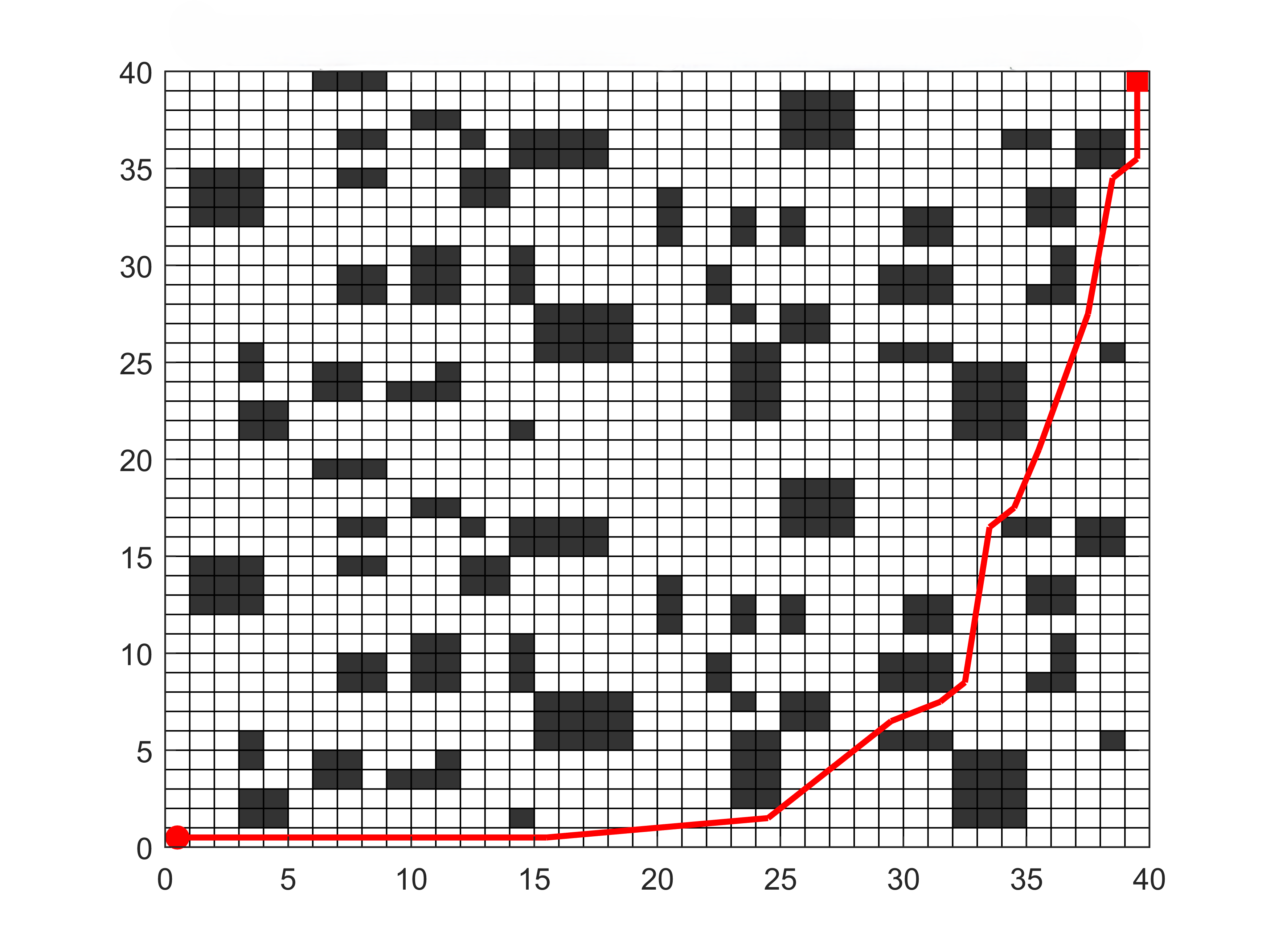}
    \caption{Robot Motion Trajectory In $40*40$ Grid}
    \label{fig:11}
\end{figure}

\begin{figure}[hbtp]
    \centering
    \includegraphics[height=45mm]{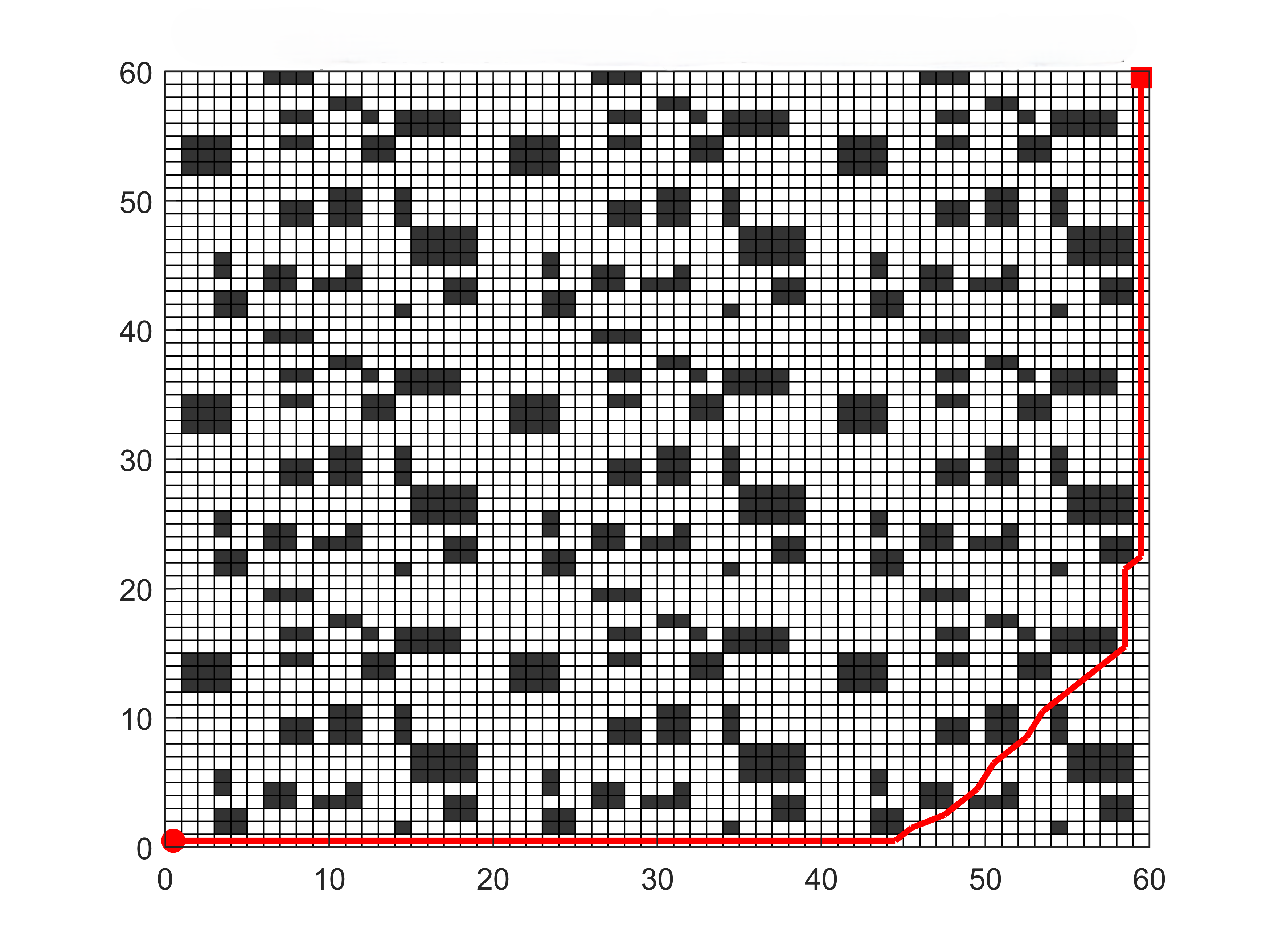}
    \caption{Robot Motion Trajectory In $60*60$ Grid}
    \label{fig:12}
\end{figure}

Tabs.\ref{table:2},\ref{table:3}, and \ref{table:4} present the comparison of path lengths generated by the MCOA and 11 other algorithms in $20$*$20$, $40$*$40$, and $60$*$60$ grid environments, respectively. Fig.\ref{fig:10}-\ref{fig:12} illustrate the motion trajectories of the robot planned by the MCOA  in the three grid environments.

In the $20$*$20$ grid environment, the MCOA plans shorter paths with fewer turning points compared to the other $11$ algorithms, achieving an average path length of $28.39$, which is 8\% lower than the average of the other algorithms. Moreover, in all experiments, MCOA consistently delivers the best performance in both optimal and worst cases.

In the $40$*$40$ grid environment, the MCOA also plans shorter paths with fewer turning points compared to the other $11$ algorithms, achieving an average path length of $62.61$, which is 48.4\% lower than the average of the other algorithms. Its optimal path length is $57.73$, significantly shorter than the others (all above $60$).

In the $60$*$60$ grid environment, the MCOA achieves the shortest average path length of $101.70$, which is 75.6\% lower than the average of the other 11 algorithms. Although GOOSE shows similar average performance to MCOA, the difference between its optimal and worst-case values is $41.95$, compared to MCOA's $18.79$, highlighting MCOA's higher stability. In terms of optimal performance, MCOA performs nearly on par with WOA.

\begin{table*}[htbp]
\centering
    \caption{$20*20$ Grid Map Results} 
    \begin{tabular}{ccccccccccccc}\\\hline 
        Metric/Algorithm & DBO & WOA & NRBO & LEA & HLOA & HEOA & GOOSE & CPO & GWO & BKA & COA & MCOA\\\hline
        Mean Value* & 29.88 & 29.51 & 30.13 & 35.28 & 29.59 & 31.32 & 30.02 & 34.34 & 29.89 & 31.30 & 29.62 & \textbf{28.39}\\
        Optimal Value & 28.57 & 28.02 & 28.53 & 30.56 & 27.84 & 29.50 & 28.53 & 31.71 & 28.84 & 29.50 & 28.42 & \textbf{27.02}\\
        Worst Value & 31.54 & 31.57 & 32.55 & 51.78 & 31.62 & 32.28 & 31.58 & 39.60 & 31.58 & 33.89 & 32.29 & \textbf{31.44}\\\hline
        
    \multicolumn{4}{l}{*All values in Tabs.\ref{table:2},\ref{table:3}, and \ref{table:4} are in grid units.} \\
    \end{tabular}
    \label{table:2}
\end{table*}

\begin{table*}[htbp]
\centering
    \caption{$40*40$ Grid Map Results} 
    \begin{tabular}{ccccccccccccc}\\\hline 
        Metric/Algorithm & DBO & WOA & NRBO & LEA & HLOA & HEOA & GOOSE & CPO & GWO & BKA & COA & MCOA\\\hline
        Mean Value & 75.10 & 68.33 & 75.93 & 214.43 & 65.88 & 73.13 & 63.61 & 530.65 & 67.97 & 73.25 & 76.85 & \textbf{62.61}\\
        Optimal Value & 67.92 & 67.37 & 72.30 & 89.31 & 60.03 & 71.66 & 60.69 & 103.63 & 62.98 & 69.66 & 74.37 & \textbf{57.73}\\
        Worst Value & 76.83 & 74.37 & 76.83 & 1600 & \textbf{69.21} & 76.83 & 89.67 & 1600 & 76.24 & 75.80 & 77.41 & 69.26\\\hline
    \end{tabular}
    \label{table:3}
\end{table*}

\begin{table*}[htbp]
\centering
    \caption{$60*60$ Grid Map Results} 
    \begin{tabular}{ccccccccccccc}\\\hline 
        Metric/Algorithm & DBO & WOA & NRBO & LEA & HLOA & HEOA & GOOSE & CPO & GWO & BKA & COA & MCOA\\\hline
        Mean Value & 179.88 & 109.19 & 1160 & 782.53 & 107.19 & 111.13 & 102.33 & 1360 & 109.35 & 397.11 & 463.30 & \textbf{101.70}\\
        Optimal Value & 108.06 & \textbf{89.41} & 179.26 & 190.96 & 92.04 & 110.62 & 90.48 & 185.63 & 90.43 & 110.64 & 154.73 & 89.78\\
        Worst Value & 259.31 & 143.06 & 3600 & 3600 & 133.69 & 111.33 & 132.43 & 3600 & 145.66 & 3600 & 3600 & \textbf{108.57}\\\hline
    \end{tabular}
    \label{table:4}
\end{table*}

\subsubsection{Exp2 - Discussions}
In mobile robot path planning experiments conducted in $20$*$20$, $40$*$40$, and $60$*$60$ grid environments, the MCOA algorithm reduced the average path length by 8\%, 48.4\%, and 75.6\%, respectively, compared to the other $11$ algorithms. Experimental results show that as environmental complexity increases, the stability of many algorithms significantly declines, while MCOA consistently performs well across grid environments of varying scales, outperforming other comparative algorithms. Furthermore, under the condition of completely avoiding obstacles, MCOA generates high-quality paths with fewer turns, further highlighting its effectiveness and robustness in optimizing path planning in 2D environments.

As environmental complexity increases, such as in the $60$*$60$ grid, the optimal performance of MCOA ($89.78$) is slightly lower than that of WOA ($89.41$), although it still performs well in terms of average and worst-case path lengths. Future work could focus on improving MCOA’s optimal performance in highly complex environments, such as those with dense obstacles or dynamic constraints.

\subsection{Validity analysis}
\textbf{Internal validity:} One potential limitation of this study lies in the selection of comparison algorithms, which affects its internal validity.
While 11 path-planning algorithms were chosen for benchmarking, they may not fully represent the breadth of state-of-the-art approaches\cite{powell2003comparing}.
The absence of certain advanced algorithms may influence the comparative results and lead to an incomplete assessment of the MCOA algorithm’s relative performance.
Expanding the benchmark set to include a wider range of algorithms would enhance the reliability of the conclusions.

\textbf{External validity:} While both the 3D UAV path planning and mobile robot path planning experiments incorporate key environmental factors, they may not fully capture the complexity of real-world scenarios.
The 3D UAV experiments consider terrain and obstacles but exclude dynamic environmental changes such as sudden weather variations and unforeseen electromagnetic interference, which could impact the algorithm’s performance in practical applications \cite{raja2012optimal}.
Similarly, the mobile robot experiments rely on grid-based environmental models, which simplify real-world terrains by omitting irregularly shaped obstacles, uneven surfaces, and dynamically moving objects\cite{zhang2021complex}.
These simplifications may lead to an overestimation of the algorithm’s effectiveness in real-world deployments, highlighting the need for further validation under more complex and dynamic conditions to ensure the algorithm’s robustness.

\section{Conclusion and Future Works}
\label{sec:conclusion}

This paper introduces MCOA, a multi-strategy enhanced COA algorithm tailored for autonomous navigation, integrating refractive opposition learning, stochastic centroid-guided exploration, and adaptive competition-based selection.
By refining population diversity, balancing global and local search, and accelerating convergence, MCOA produces more optimal paths with shorter computation.
Comparative evaluations against 11 baseline algorithms demonstrate that, while adhering to real-life constraints, MCOA reduces total path cost by 16.7\% and shortens computation time by 69.2\% in 3D UAV scenarios.
In 2D mobile robot scenarios, it achieves an average 44\% reduction in path length.
These results highlight MCOA as a promising solution for real-world autonomous navigation applications  


Future research will focus on extending MCOA’s capabilities to more complex path-planning scenarios, particularly in real-time dynamic environments and higher-dimensional decision spaces.
A key focus will be enhancing MCOA’s responsiveness to real-time dynamic path planning, where external factors such as moving obstacles and environmental variations demand rapid adjustments without requiring full re-optimization.
Additionally, it's worth to explore MOCA's potential in higher-dimensional decision-making problems, where path planning must account for interdependent factors beyond spatial coordinates, including temporal constraints, velocity profiles, energy consumption, and multi-agent coordination.
To support these advancements, further improvements in computational efficiency and robustness will be investigated to enable practical deployment in real-world autonomous Navigation.

\bibliographystyle{IEEEtran}
\bibliography{main}

\end{document}